\RequirePackage{fix-cm}
\documentclass[smallextended]{svjour3}       
\smartqed  
\usepackage{graphicx}
\journalname{Autonomous Agents and Multi-Agent Systems}

\usepackage[sorting=none]{biblatex}
\bibliography{main}

\usepackage{mathtools} 
\usepackage{booktabs} 
\usepackage{tikz} 

\usepackage{times}
\usepackage{soul}
\usepackage{url}
\usepackage[hidelinks]{hyperref}
\usepackage[utf8]{inputenc}
\usepackage[small]{caption}
\usepackage{graphicx}
\usepackage{amsmath}
\usepackage{amssymb}

\usepackage{booktabs}
\usepackage{algorithm}
\usepackage{algorithmic}
\usepackage{subcaption}
\usepackage{svg}

\begin{document}

\title{MACRPO: Multi-Agent Cooperative Recurrent Policy Optimization
}


\author{Eshagh Kargar*        \and
        Ville Kyrki 
}


\institute{Eshagh Kargar \at
              Aalto University, Finland \\
              \email{eshagh.kargar@aalto.fi}           
           \and
           Ville Kyrki \at
              Aalto University, Finland \\
              \email{ville.kyrki@aalto.fi}
}

\date{Received: date / Accepted: date}

\maketitle

\begin{abstract}
    This work considers the problem of learning cooperative policies in multi-agent settings with partially observable and non-stationary environments without a communication channel. 
    We focus on improving information sharing between agents and propose a new multi-agent actor-critic method called \textit{Multi-Agent Cooperative Recurrent Proximal Policy Optimization} (MACRPO).
    We propose two novel ways of integrating information across agents and time in MACRPO: 
    First, we use a recurrent layer in critic's network architecture and propose a new framework to use a meta-trajectory to train the recurrent layer. 
    This allows the network to learn the cooperation and dynamics of interactions between agents, and also handle partial observability. 
    Second, we propose a new advantage function that incorporates other agents' rewards and value functions.
    We evaluate our algorithm on three challenging multi-agent environments with continuous and discrete action spaces, Deepdrive-Zero, Multi-Walker, and Particle environment. We compare the results with several ablations and state-of-the-art multi-agent algorithms such as QMIX and MADDPG and also single-agent methods with shared parameters between agents such as IMPALA and APEX. 
    The results show superior performance against other algorithms. The code is available online at \url{https://github.com/kargarisaac/macrpo}.
\keywords{Multi-Agent \and Reinforcement Learning \and Cooperative}
\end{abstract}

\section{Introduction}
\label{intro}

\begin{figure*}[h]
    \includegraphics[width=\textwidth]{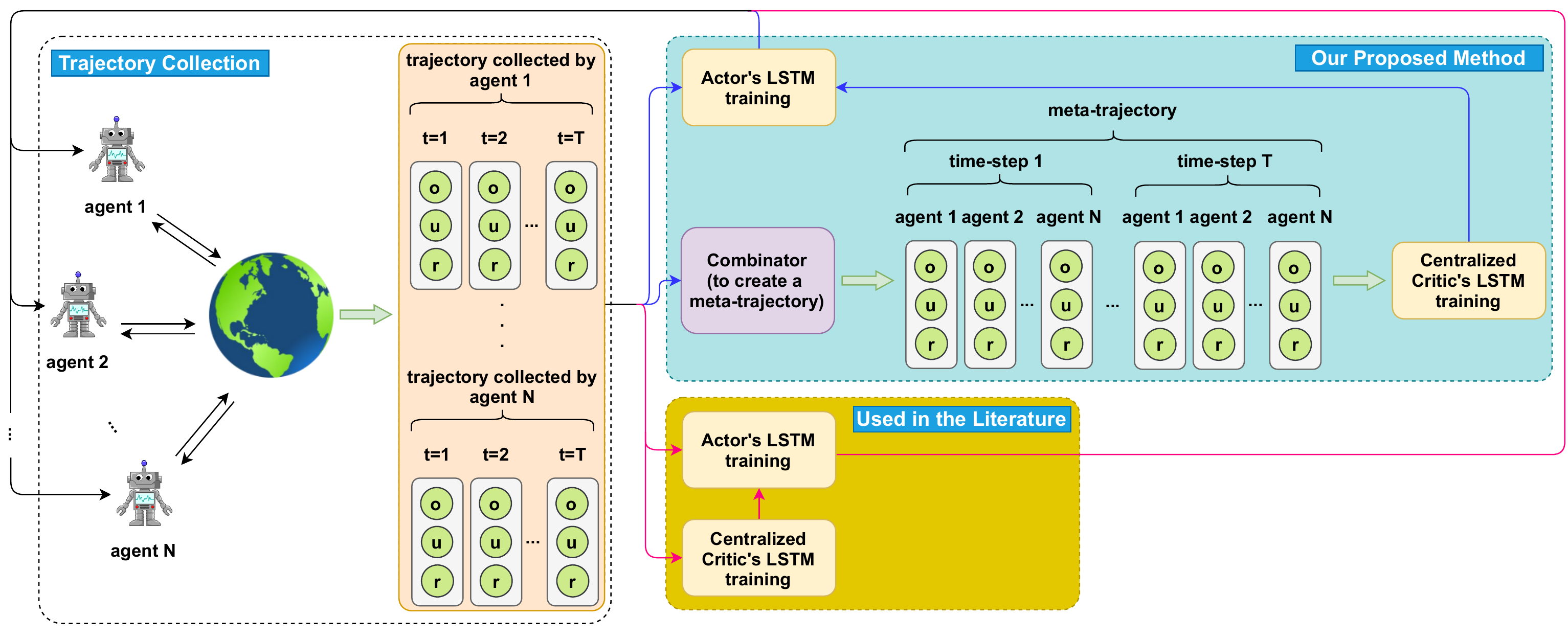}
    \centering
    \caption{Different frameworks for information sharing. Our proposed method and the standard approach for information sharing through agents are shown in separate boxes. Blue arrows are for ours, and the red ones are for the standard approach to share parameters. After collecting trajectories by agents, ours, in addition to sharing parameters between agents, uses the meta-trajectory to train the critic's LSTM layer. This allows the critic to learn the interaction between agents along the trajectories through its hidden state. In contrast, the literature approach, which does parameter sharing, uses separate trajectories collected by agents to train the LSTM layer. For more details about the network architectures, please see Fig~\ref{fig:actor_critic}.}
    \label{fig:lstm_training1}
\end{figure*}

While reinforcement learning (RL) has gained popularity in policy learning, many problems which require coordination and interaction between multiple agents cannot be formulated as single-agent reinforcement learning. Examples of such scenarios include self-driving cars~\cite{shalev2016safe}, multiplayer games~\cite{berner2019dota,vinyals2019grandmaster}, and distributed logistics~\cite{ying2005multi}.
Solving these kind of problems using single-agent RL is problematic, because the interaction between agents and the non-stationary nature of the environment due to multiple learning agents can not be considered~\cite{hernandez2019survey}.
Multi-agent reinforcement learning (MARL) and cooperative learning between several interacting agents can be beneficial in such domains and has been extensively studied~\cite{nguyen2020deep,hernandez2019survey}. 
However, when several agents are interacting with each other in an environment without real-time communication, the lack of communication deteriorates policy learning. 
In order to alleviate this problem, we propose to share information during training to learn a policy that implicitly considers other agents’ intentions to interact with them in a cooperative manner.
For example, in applications like autonomous driving and in an intersection, knowing about other cars' intentions can improve the performance, safety, and collaboration between agents. 

A standard paradigm for multi-agent planning is to use the centralized training and decentralized execution approach~\cite{kraemer2016multi,foerster2016learning,lowe2017multi,foerster2017counterfactual}, also taken in this work. 

In this work, we propose a new cooperative multi-agent reinforcement learning algorithm, which is an extension to Proximal Policy Optimization (PPO), called \textit{Multi-Agent Cooperative Recurrent Proximal Policy Optimization} (MACRPO). 
MACRPO combines and shares information across multiple agents in two ways: First, in network architecture using long short term memory (LSTM) layer and train it by creating a meta-trajectory from trajectories collected by agents, as shown in Fig~\ref{fig:lstm_training1}. This allows the critic to learn the cooperation and dynamics of interactions between agents, and also handle the partial observability. Second, in the advantage function estimator by considering other agents' rewards and value functions. 

MACRPO uses a centralized training and decentralized execution paradigm that the centralized critic network uses extra information in the training phase and switches between agents sequentially to predict the value of a state for each agent. In the execution time, only the actor networks are used, and each learned policy (actor network) only uses its local information (i.e., its observation) and acts in a decentralized manner.

Moreover, in environments with multiple agents that are learning simultaneously during training, each agent's policy and the dynamics of the environment, from each agent's perspective, is constantly changing. This causes the non-stationarity problem~\cite{hernandez2019survey}. To reduce this effect, MACRPO uses an on-policy approach and the most recent collected data from the environment.

In summary, our contributions are as follows: (1) proposing a cooperative on-policy centralized training and decentralized execution framework that is applicable for both discrete and continuous action spaces; (2) sharing information across agents using two ways: a recurrent component in the network architecture which uses a combination of trajectories collected by all agents and an advantage function estimator that uses a weighted combination of rewards and value functions of individual agents; (3) evaluating the method on three cooperative multi-agent tasks: DeepDrive-Zero~\cite{craig_quiter_2020_3871907}, Multi-Walker~\cite{gupta2017cooperative}, and Particle~\cite{mordatch2017emergence} environments, demonstrating similar or superior performance compared to the state-of-the-art. 

The rest of this paper is organized as follows. 
The review of related works in Section \ref{related_works} demonstrates that while MARL has been extensively studied, existing approaches do not address the dynamics of interaction between agents in detail. 
In Section \ref{background}, we provide the required background in Markov Games and Proximal Policy Optimization. 
The problem definition and the proposed method are described in Section \ref{method}, with emphasis on the two innovations, meta-trajectory for recurrent network training and joint advantage function. 
Then, Section \ref{experiments} presents empirical evaluation in three multi-agent environments showing superior performance of the proposed approach compared to state-of-the-art. 
Finally, in Section \ref{conclusion} we conclude that implicit information sharing can be used to improve cooperation between agents while discussing its limitations in settings with high number of agents. 

\section{RELATED WORK}
\label{related_works}
The most straightforward and maybe the most popular approach to solve multi-agent tasks is to use single-agent RL and consider several independent learning agents. Some prior works compared the performance of cooperative agents to independent agents, and tried independent Q-learning~\cite{tan1993multi} and PPO with LSTM layer~\cite{bansal2017emergent}, but they did not work well in practice~\cite{matignon2012independent}. Also, \cite{zhao2020research} tried to learn a joint value function for two agents and used PPO with LSTM layer to improve the performance in multi-agent setting.

In order to use single-agent RL methods for multi-agent setting, improve the performance, and speed up the learning procedure, some works used parameter sharing between agents~\cite{gupta2017cooperative,terry2020parameter}. Especially in self-play games, it is common to use the current or older versions of the policy for other agents~\cite{berner2019dota}. We will compare our proposed method with several state-of-the-art single-agent RL approaches with shared parameters between agents proposed in~\cite{terry2020parameter} in the experiments section.
Our way of training the LSTM layer in the critic differs from parameter sharing used in the literature such that instead of using separate LSTMs for each agent, the LSTM layer in our method has a shared hidden state, which is updated using a combination of all agents' information. This lets the LSTM layer to learn about the dynamics of interaction and cooperation between agents and across time.

In addition to using single-agent RL methods with or without parameter sharing, some other works focused on designing multi-agent RL algorithms for multi-agent settings. 
In multi-agent environments, considering communication between agents will lead to designing multi-agent methods. 
The communication channel is often limited, leading to methods that try to optimize the communication including message structure \cite{mao2020learning, kullu2017acmics}. 
However, in some environments, there is no explicit communication channel between agents. For example, consider an autonomous driving environment without connection between cars. Finding a solution to address this problem and decrease the lack of communication effect seems necessary.

A recently popularized paradigm to share information between agents is to use centralized training and decentralized execution. 
In general, we can categorize these types of approaches into two groups: value-based and actor-critic-based. In value-based methods, the idea is to train a centralized value function and then extract the value functions for each agent from that to act in a decentralized manner in the execution time~\cite{sunehag2018value,rashid2018qmix}. 
On the other hand, the actor-critic-based methods have actor and critic networks~\cite{lowe2017multi,foerster2017counterfactual}. The critic network has access to data from all agents and is trained in a centralized way, but the actors have only access to their local information. They can act independently in the execution time. The actors can be independent with individual weights ~\cite{lowe2017multi} or share the policy with shared weights~\cite{foerster2017counterfactual}. In this work, we use an actor-critic-based method with centralized training and decentralized execution, providing two innovations to improve information sharing without communication channel between agents during execution. 

In ~\cite{foerster2017counterfactual}, which is one of the works near ours, the actor is recurrent, but the critic is a feed-forward network, whereas our actor and critic are both recurrent, and the recurrent layer in our critic has a crucial role in our method. Their method is also for settings with discrete action spaces, whereas we test our method on three environments with both discrete and continuous action spaces. 

The other similar work to ours, which is one of the most popular MARL methods, is the multi-agent deep deterministic policy gradient (MADDPG)~\cite{lowe2017multi} that proposed similar frameworks with centralized training and decentralized execution. They tested their method on some Particle environments~\cite{mordatch2017emergence}. 
Their approach differs from ours in the following ways: (1) They do not have the LSTM (memory) layer in their network, whereas the LSTM layer in the critic network plays a critical role in our method. It helps to learn the interaction and cooperation between agents and also mitigate the partial observability problem. (2) They tested their method on environments with discrete action spaces, but we test our method on environments with both continuous and discrete action spaces. (3) They consider separate critic networks for each agent, which is beneficial for competitive scenarios, whereas we use a single critic network and consider the cooperative tasks. (4) Their method is off-policy with replay buffer, and they combat the non-stationarity problem by centralized training. In contrast, our approach, in addition to centralized training, is an on-policy method without replay buffer allowing the networks to use the most recent data from the environment. 
We will compare our method with MADDPG and show that ours has comparable or superior performance. \cite{wang2020r} extends the MADDPG idea and adds a recurrent layer into the networks, but they have separate actors and critics for agents, similar to MADDPG, and recurrent hidden states of critics are isolated, and there is no combination of information in them. They also tested their method on one environment with a discrete action space.


\section{BACKGROUND}
\label{background}
\subsection{MARKOV GAMES}
In this work, we consider a multi-agent extension of Partially Observable Markov Decision Processes (MPOMDPs) which is also called partially observable Markov games~\cite{littman1994markov}. A Markov game for \textit{N} agents is defined by a set of states $ \mathcal{S} $ describing the possible configurations of all agents, a set of actions $ \mathcal{U}_1,\ldots, \mathcal{U}_N$ and a set of observations $ \mathcal{O}_1, \ldots , \mathcal{O}_N$ for each agent. A transition function $ \mathcal{T}: \mathcal{S} \times \mathcal{U}_1 \times \ldots \times \mathcal{U}_N \rightarrow \mathcal{S}$ gives the probability distribution of the next state as a function of current state and actions. Each agent \textit{i} uses a stochastic policy $\pi_{\theta_i}: \mathcal{O}_i \times \mathcal{U}_i \rightarrow [0, 1] $ to choose action and go to the next state based on the transition function and achieve reward $r_i : \mathcal{S} \times \mathcal{U}_i \rightarrow \mathcal{R} $. We consider games where the reward can be decomposed into individual agent rewards $r_i$.
Each agent \textit{i} aims to maximize the rewards for all agents in a cooperative way~\cite{lowe2017multi}. 

\subsection{Proximal Policy Optimization} 
Proximal Policy Optimization (PPO) is a family of policy gradient methods for solving reinforcement learning problems, which alternate between sampling data through interaction with the environment, and optimizing a surrogate objective function using stochastic gradient descent while ensuring the deviation from the policy used to collect the data is relatively small. One of the differences between PPO and standard policy gradients is that standard policy gradient methods perform one gradient update per data sample, while PPO does multiple epochs of minibatch updates. The main objective that PPO tries to maximize is 
\begin{align*}
    L^{CLIP}(\theta) = \hat{ \mathop{\mathbb{E}} }_t [min(f_t(\theta) \hat{A}_t, clip(f_t(\theta), 1 - \epsilon, 1 + \epsilon) \hat{A}_t)] 
\end{align*}

where $\hat{A}_t$ is the Generalized Advantage Estimation (GAE), $ \epsilon $ is a hyperparameter and $ f_t(\theta) $ denotes the probability ratio $ f_t(\theta) = \frac{ \pi_{\theta}(u_t | o_t) }{\pi_{{\theta}_{old}}(u_t | o_t)}$. This objective prevents large policy updates by clipping the ratio between $ [ 1 - \epsilon, 1 + \epsilon ] $.   

The total objective function for PPO is
\begin{align*}\label{eq:L_ppo_mixed}
    L^{CLIP+VF+S}_t (\theta) = \hat{ \mathop{\mathbb{E}} }_t [ L^{CLIP}_t (\theta) - c_1 L^{VF}_t (\theta) + c_2 S[\pi_{\theta}](o_t)  ]
\end{align*}
where $c_1$, $c_2$ are coefficients, $S$ denotes an entropy bonus, and $ L^{VF}_t $ is a squared-error loss for critic network:
\begin{equation}\label{eq:value_loss}
    L^{VF}_t (\theta)  = (V_{\theta}(o_t) - V^{targ}_t)^2
\end{equation}
In the above equations, $V_{\theta}(o_t)$ is the state-value function and $ \theta $ denotes the combined parameter vector of actor and critic networks~\cite{schulman2017proximal}.

\section{Method}
\label{method}
In this section, we first explain the problem setting and review the proposed ideas to solve it. We then proceed with describing the proposed framework used in MACRPO for training the LSTM layer. Finally, we present a detailed description of MACRPO objective function including the novel advantage estimation approach and the learning algorithm.

\subsection{Problem Setting and Solution Overview}

Information sharing across agents can help to improve the performance and speed up the learning procedure~\cite{gupta2017cooperative,foerster2017counterfactual,terry2020parameter}. 
In this work, we focus on improving information sharing between agents in multi-agent settings in addition to just sharing parameters across actors. 
We propose Multi-Agent Cooperative Recurrent Proximal Policy Optimization (MACRPO) algorithm, which is a multi-agent cooperative algorithm and uses the centralized learning and decentralized execution framework. 
In order to improve information sharing between agents, MACRPO, in addition to use parameter sharing, uses two novel ideas: (a) a recurrent component in the critic architecture which uses a meta-trajectory, created by a combination of trajectories collected by all agents, to train and is described in Section~\ref{MACRPO_framework}, (b) an advantage function estimator that uses a weighted combination of rewards and value functions of individual agents, which is explained in Section~\ref{MACRPO_obj}.

We consider the case where each agent has its local reward and wants to cooperate with other agents to solve a task. In other words, each agent wants to maximize its future cumulative reward while considering maximizing all agents' total reward together in a collaborative way. This paper does not solve the credit assignment problem in multi-agent cooperative games, where all agents in a team have the same team reward. This algorithm can be applied to that kind of problem, but it is not optimized for that setting.

\subsection{MACRPO Framework}\label{MACRPO_framework}

Fig~\ref{fig:lstm_training1} shows the proposed framework for training MACRPO's actor and critic. This framework has one critic network and one actor network. To consider the partial observability of multi-agent settings, we use LSTM layers in the neural network architectures, both in actor and critic networks. 

The actor is trained using trajectories collected by all agent. We denote the shared weights of actors with $\theta_a$ and use the same and the latest trained weights for all agents. The agents will behave differently in the execution time because of different inputs. 

The trajectory data for one episode with length \textit{T} for agent \textit{i}, which is used for training actor is denoted as
\begin{align*}
    o^i_1, u^i_1, r^i_1, \ldots, o^i_T, u^i_T, r^i_T
\end{align*}

To allow the critic network to integrate information across agents and across time, we use all agents' trajectories in one roll-out and puts them in a sequence to create a meta-trajectory, and train its network using that— first data from agent 1, then agent 2, then agent 3, and so on. See Fig~\ref{fig:lstm_training1}. We denote the critic's weights with $ \theta_c $. 
To remove the dependency to agents' orders, we randomize agents' orders in environments with more than two agents. 
So the LSTM layer in each time-step, which we have data from N agents, will switch between all agents and get their observation, or any other additional input that we want to feed into the critic, such as the previous action of that agent.
The meta-trajectory for training the critic network is
    \begin{align*}
        (o^1_1, \ldots, o^N_1), (u^1_1, \ldots, u^N_1), (r^1_1, \ldots, r^N_1), \ldots \\
        , (o^1_T, \ldots, o^N_T), (u^1_T, \ldots, u^N_T), (r^1_T, \ldots, r^N_T)
    \end{align*}

\begin{figure}[t!]
    \begin{subfigure}{0.5\textwidth}
        \centering
        \includegraphics[width=0.73\linewidth]{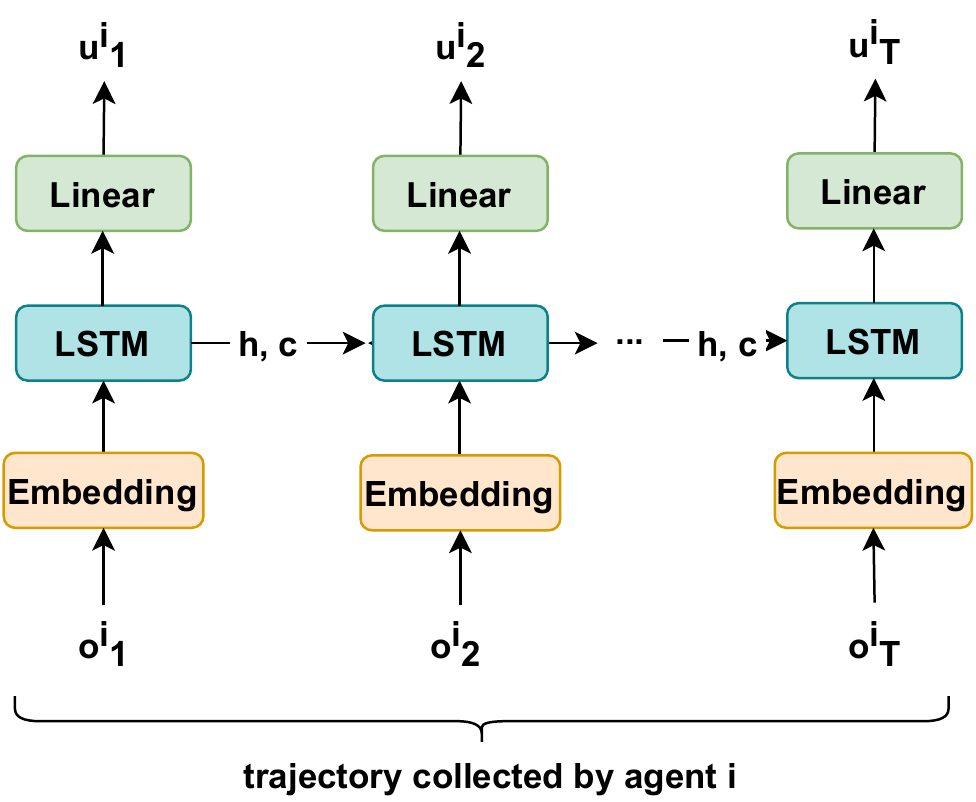}  
        \caption{}
        \label{fig:actor}   
    \end{subfigure}
    \begin{subfigure}{.5\textwidth}
        \centering
        \includegraphics[width=1\linewidth]{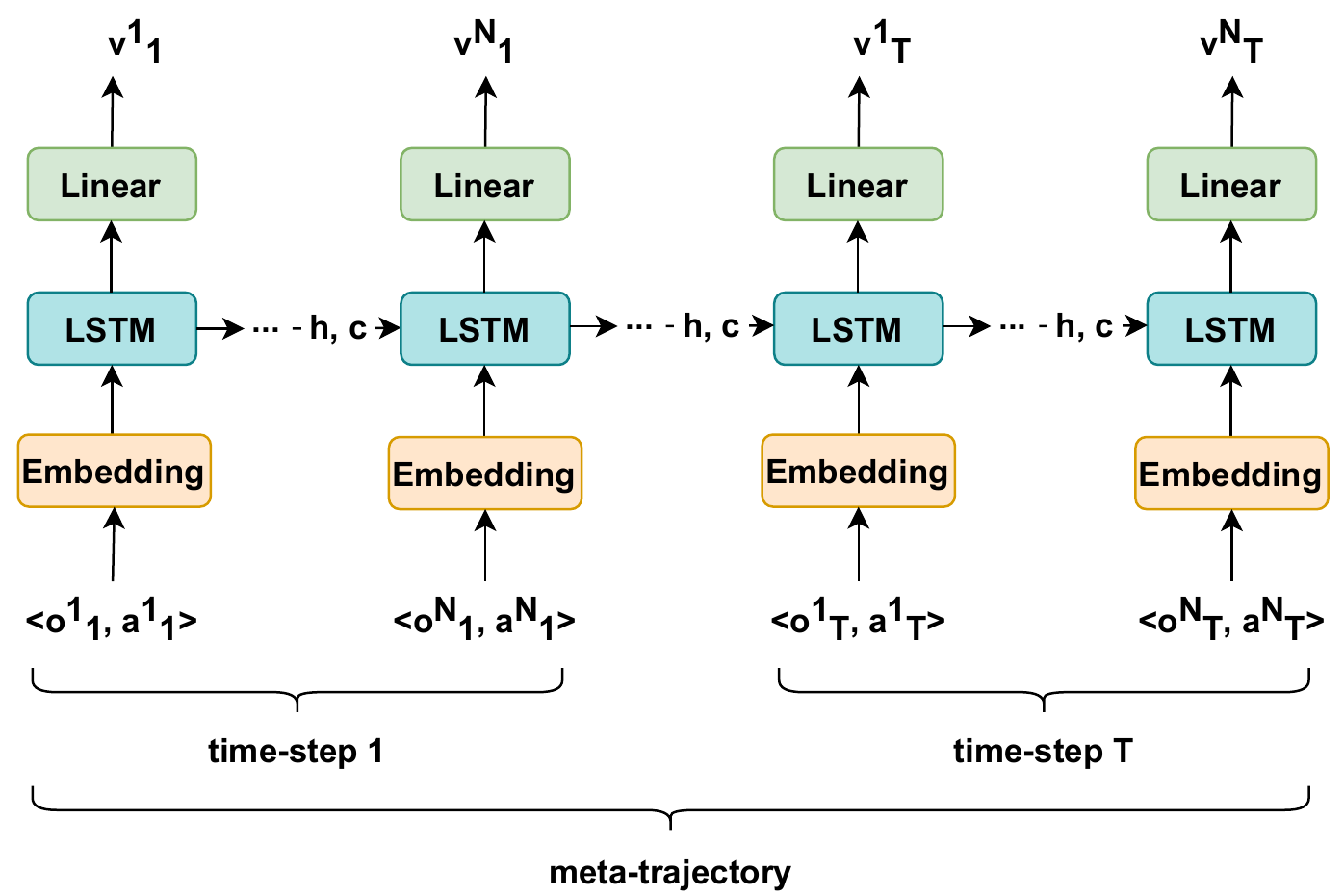}  
        \caption{}
        \label{fig:critic}
    \end{subfigure}
    \caption{Actor and critic network architectures. (a) Actor network architecture for agent \textit{i} which uses the collected trajectory by itself, (b) The centralized critic network architecture which uses the created meta-trajectory. Note that $u$ and $v$ denote action and value, respectively.}
    \label{fig:actor_critic}
\end{figure}

By using the above meta-trajectory, the critic network can use the data from all agents and learn about the agents' history, the interaction between agents, and somehow the environment's dynamic using its hidden state. 
In other words, MACRPO is able to consider temporal dynamics using the LSTM layer, which incorporates a history of states and actions across all agents. Modeling temporal dynamics allows the latent space to model differential quantities such as the rate of change (derivative) between the distance of two agents and integral quantities such as the running average of the distance.

Additionally, the hidden state of recurrent networks can be viewed as a communication channel that allows information to flow between agents to create richer training signals for actors during training. The network will update the hidden state in each time-step by getting the previous hidden state and the data from the agent \textit{i} in that time-step. The network architectures for actor and critic are shown in Fig~\ref{fig:actor_critic}.

\subsection{Objective Function}\label{MACRPO_obj}

In addition to the LSTM layer, we propose a novel advantage function estimator and discounted return equations to integrate information across all agents. We consider the $ V^{targ}_t $ in Equation~(\ref{eq:value_loss}) as discounted return and calculate it for agent \textit{i} at time \textit{t} as
\begin{equation}\label{eq:ma_discounted_return}
    R_t^{i} = \overline{r}_t + \gamma \overline{r}_{t+1} + \ldots + \gamma^{T-t+1} \overline{V}(o_T^i)
\end{equation}
where
\begin{equation}\label{eq:ma_mean_r}
    \medmuskip=2mu   
    \thickmuskip=3mu 
    \renewcommand\arraystretch{1.5}
    \overline{r}_t = \frac{r^i_t + \beta \sum_{j \neq i} r^j_t}{N} , \quad  \overline{V}(o_T^i) = \frac{V(o^i_T) + \beta \sum_{j \neq i} V(o^j_T)}{N}
\end{equation}
where $ r_t^i $ is the reward value for agent \textit{i} at time \textit{t}, $ \gamma $  is the discount factor, $ \beta $ is a weight for other agents data to calculate a weighted mean, and $ V(o^i_T) $ is the value for the final state of agent \textit{i}.

Also, the advantage for each agent \textit{i} is calculated as
\begin{equation}\label{eq:ma_advantage}
    \hat{A}_t^{i} = \delta_t^{i} + (\gamma \lambda) \delta_{t+1}^{i} + \ldots + \ldots + (\gamma \lambda)^{T-t+1} \delta_{T-1}^{i}
\end{equation}
where
\begin{align}\label{eq:ma_delta}
        \delta_t^{i} = & \frac{1}{N}  [ r_t^{i} + \gamma V(o_{t+1}^{i}) - V(o_t^{i}) + \nonumber\\
        + & \beta \sum_{j \neq i} (r_t^{j} + \gamma V(o_{t+1}^{j}) - V(o_t^{j}))]
\end{align}
where $ \lambda $ is for GAE algorithm, and $ V(o_t^i) $ is the state-value at time \textit{t} for agent \textit{i}. The intuition behind considering other agents' information is, that agent \textit{i} by doing action $u_t^i$ affects the rewards and values of the other agents in addition to its own. 

It is worth mentioning that in MACRPO, we use separate networks for actor and critic. Therefore, the objective functions of the actor and critic networks are separate.

\begin{algorithm}[t]
\caption{MACRPO}
\label{alg:macpp_algo}
\begin{algorithmic}[1] 
\STATE Randomly initialize actor and critic networks' parameters \(\theta_c\) and \(\theta_a \)
\FOR{iteration=1, 2, ...}
    \FOR{environment=1, 2, ..., E}
        \STATE Run all N agents with latest trained weights in the environment for T time-steps and collect data
        \STATE Combine collected trajectories by all agents according to Fig~\ref{fig:lstm_training1}
        \STATE Compute discounted returns and advantage estimates using Equations~(\ref{eq:ma_advantage}, \ref{eq:ma_discounted_return})
    \ENDFOR
    \FOR{epoch=1, ..., K}
        \FOR{minibatch=1, ..., M}
            \STATE Calculate the loss functions using Equations~(\ref{eq:ma_policy_loss}, \ref{eq:ma_value_loss})
            \STATE Update Actor and Critic parameters via Adam
        \ENDFOR
    \ENDFOR
\ENDFOR
\end{algorithmic}
\end{algorithm}

The actor's objective function in the shared weights case is defined as
\begin{equation}\label{eq:ma_policy_loss}
    L^{CLIP+S}_t (\theta_a) = \hat{ \mathop{\mathbb{E}} }_t [ L^{CLIP}_t (\theta_a) + c S[\pi_{\theta_a}](o_t)  ]
\end{equation}
and the critic objective function is
\begin{equation}\label{eq:ma_value_loss}
    L^{VF}_t (\theta)  = (V_{\theta_c}(o_t) - V^{targ}_t)^2 .
\end{equation}

The MACRPO algorithm with parallelized implementation is shown in Algorithm~\ref{alg:macpp_algo}.


\section{EXPERIMENTS}
\label{experiments}

This section presents empirical results that compare the performance of our proposed method, MACRPO, with several ablations to see the effect of each proposed novelty. We also compare our method with recent advanced RL methods in both single-agent domain with shared parameters between agents~\cite{gupta2017cooperative,terry2020parameter} and multi-agent domain like MADDPG~\cite{lowe2017multi} and QMIX~\cite{rashid2018qmix}.

\subsection{Test Environments}

We test our method in three MARL environments. In two of them, DeepDrive-Zero~\cite{craig_quiter_2020_3871907} and Multi-Walker~\cite{pettingZoo2020} environments, the action space is continuous, and in the third environment, the Particle environment~\cite{mordatch2017emergence}, the action space is discrete. 

\begin{figure*}[h!]
    \begin{subfigure}{0.33\textwidth}
        \includegraphics[width=0.67\linewidth]{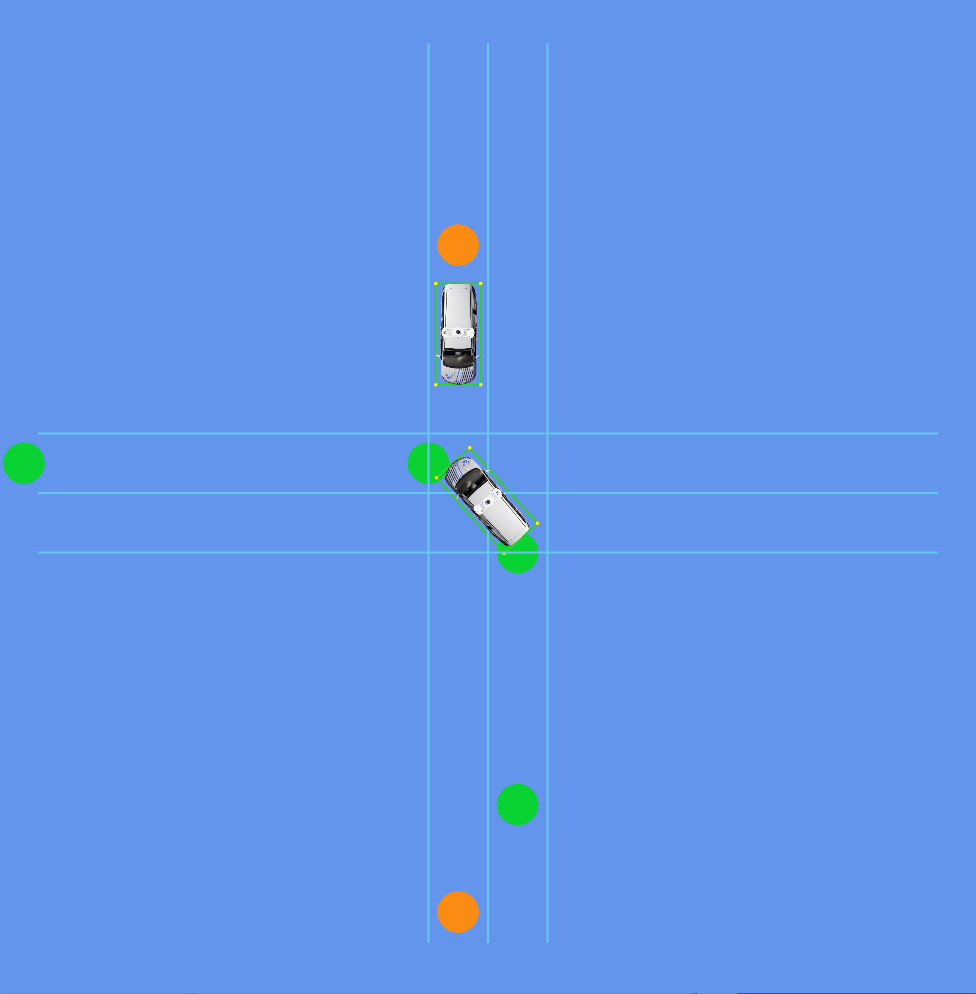}
        \centering
        \caption{}
        \label{fig:dd0_env}  
    \end{subfigure}
    \begin{subfigure}{.33\textwidth}
        \includegraphics[width=1.05\linewidth]{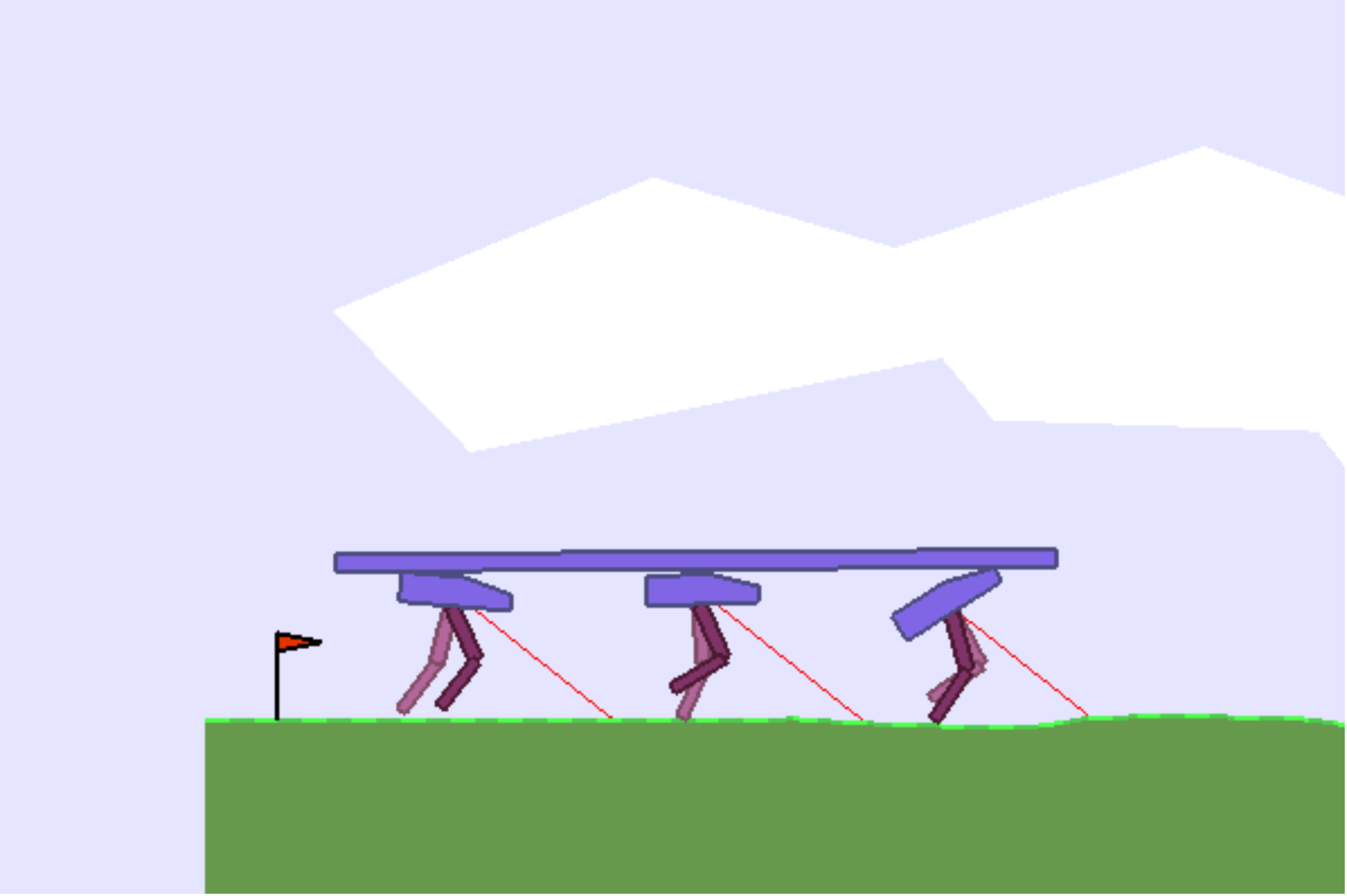}
        \centering
        \caption{}
        \label{fig:mw_env}
    \end{subfigure}
    \begin{subfigure}{.33\textwidth}
        \includegraphics[width=0.79\linewidth]{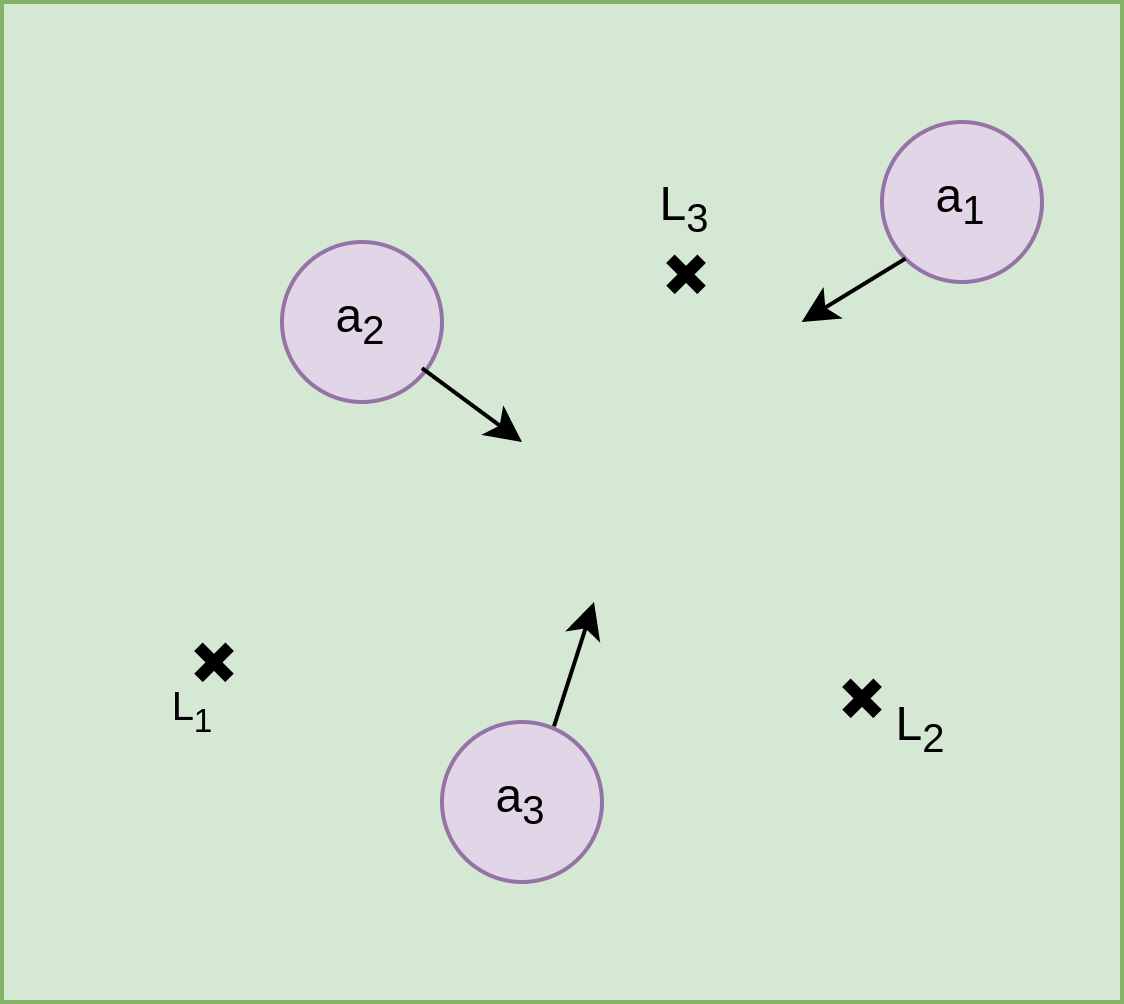}
        \centering
        \caption{}
        \label{fig:particle_env}
    \end{subfigure}
    \caption{Considered MARL simulation environments (a) DeepDrive-Zero environment: an unprotected left turn scenario, (b) Multi-Walker environment, (c) Particle environment: cooperative navigation.}
    \label{fig:test_envs}
\end{figure*}

\paragraph{DeepDrive-Zero Environment:}
DeepDrive-Zero environment~\cite{craig_quiter_2020_3871907} is a 2D  simulation environment for self-driving cars which uses a bike model for the cars. We use the unsignalized intersection scenario in this work, which is shown in Fig~\ref{fig:dd0_env}. To test our algorithm, we consider two cars in the environment, one starts from the south and wants to follow the green waypoints to do an unprotected left-turn, and the other one starts from the north and wants to go to the south and follow the orange waypoints. The agents need to learn to cooperate and negotiate to reach their destination without any collision.

The observation space is a vector with continuous values. Each agent in the environment receives some information about itself, as well as information from other agents. This information can come from some modules like Perception, Localization, and HDMap in a self-driving car and be used by the decision making and control modules. The observation vector for each agent contains some information about the agent itself like distance and angle to waypoints, velocity, acceleration, and distance to left and right lanes, and also some information about the other agents like the relative velocity of the other agent to the ego agent, velocity and acceleration of the other car, angles to corners of the other agent, and distance to corners of the other agent.

Each action vector element is continuous from -1 to 1: steering, acceleration, and braking. Negative acceleration can be used to reverse the car, and the network outputs are scaled to reflect physically realistic values. This environment also has a discretized version that we used in discrete action methods.

The reward function is a weighted sum of several terms like speed, reaching the destination, collision, G-force, jerk, steering angle change, acceleration change, and staying in the lane. Initially, we used $0.5$, $1$, $4$, $1 \times 10^{-7}$, $6 \times 10^{-6}$, $0.0001$, $0.0001$, $0.001$ as weights, then used curriculum learning to smooth the driving behavior. 

\paragraph{Multi-Walker Environment:}
The multi-walker environment is a multi-agent continuous control locomotion task introduced in ~\cite{gupta2017cooperative}. The environment contains agents (bipedal walkers) that can actuate the joints in each of their legs and convey objects on top of them. Fig~\ref{fig:mw_env} shows a snapshot from the environment. To keep the package balanced and move it as far to the right as possible, the walkers must coordinate their movements. A positive reward is given to each walker locally, based on the change in the package distance summed with 130 times the change in the walker’s position. A walker is given a reward of -100 if they fall, and all walkers receive a reward of -100 if the package falls while moving forward has a reward of 1. By default, the environment is done whenever a walker or package falls or when the walkers reach the edge of the terrain. The action space is continuous, with four values for torques applied to each walker's leg. The observation vector for each walker is a 32-dimensional vector that contains information about nearby walkers as well as data from some noisy LiDAR sensors.

\paragraph{Cooperative Navigation in Particle Environment:}
Using the particle environment package from OpenAI~\cite{lowe2017multi}, we created a new environment based on the cooperative navigation environment. This new environment consists of N agents and N landmarks, and agents must avoid collisions and cooperate to reach and cover all landmarks. We assign each agent a landmark and calculate its local reward based on its proximity to its landmark and collisions with other agents. As a result, agents will have different reward values; not one shared reward. Each agent's observation data is its position and velocity, as well as the relative position of other agents and landmarks. There are five discrete actions in the action space: up, down, left, right, and no move. After 25 time-steps, the episode ends. Fig~\ref{fig:particle_env} shows the simulation environment.

\subsection{Ablation Study}

Four ablations were designed to evaluate each novelty. In all cases, the parameter sharing proposed in~\cite{gupta2017cooperative,terry2020parameter} was used:

\paragraph{FF-NIC} \textit{(Feed-forward multi-layer perceptron (MLP) network + no information combination)}: two feed-forward neural networks for actor and critic. The GAE is calculated using the single-agent PPO GAE equation~\cite{schulman2017proximal}.

\paragraph{FF-ICA} \textit{(Feed-forward MLP network + information combination using the advantage estimation function)}: This case is similar to the previous case, but the GAE is calculated using Equation~(\ref{eq:ma_advantage}).
    
\paragraph{LSTM-NIC} \textit{(LSTM network + no information combination)}: two networks with LSTM layers for actor and critic. There is no information sharing between agents through GAE calculation or the LSTM's hidden state. The GAE is calculated using the single-agent PPO GAE equation~\cite{schulman2017proximal}.

\paragraph{LSTM-ICA} \textit{(LSTM network + information combination using the advantage estimation function but not through the LSTM layer)}: This case is identical to the previous case, but the GAE is calculated using Equation~(\ref{eq:ma_advantage}).

\paragraph{LSTM-ICF} \textit{(LSTM network + information sharing using both the advantage estimation function and an LSTM layer in the critic network (full proposed method))}: two networks with LSTM layers for actor and critic. In addition to parameter sharing between actors, the information integration is done through both the advantage estimation function and the LSTM's hidden state in the centralized critic network, shown in Fig ~\ref{fig:lstm_training1}.

Also, in order to see the effect of the \(\beta \) value in Equations~(\ref{eq:ma_mean_r}, \ref{eq:ma_delta}), the proposed method was evaluated with different \(\beta \) values.

All experiments were repeated with identical random seeds for each method to reduce the effect of randomness.

\begin{figure}[t!]
    \begin{subfigure}{\textwidth}
        \centering
        \includegraphics[width=0.8\linewidth]{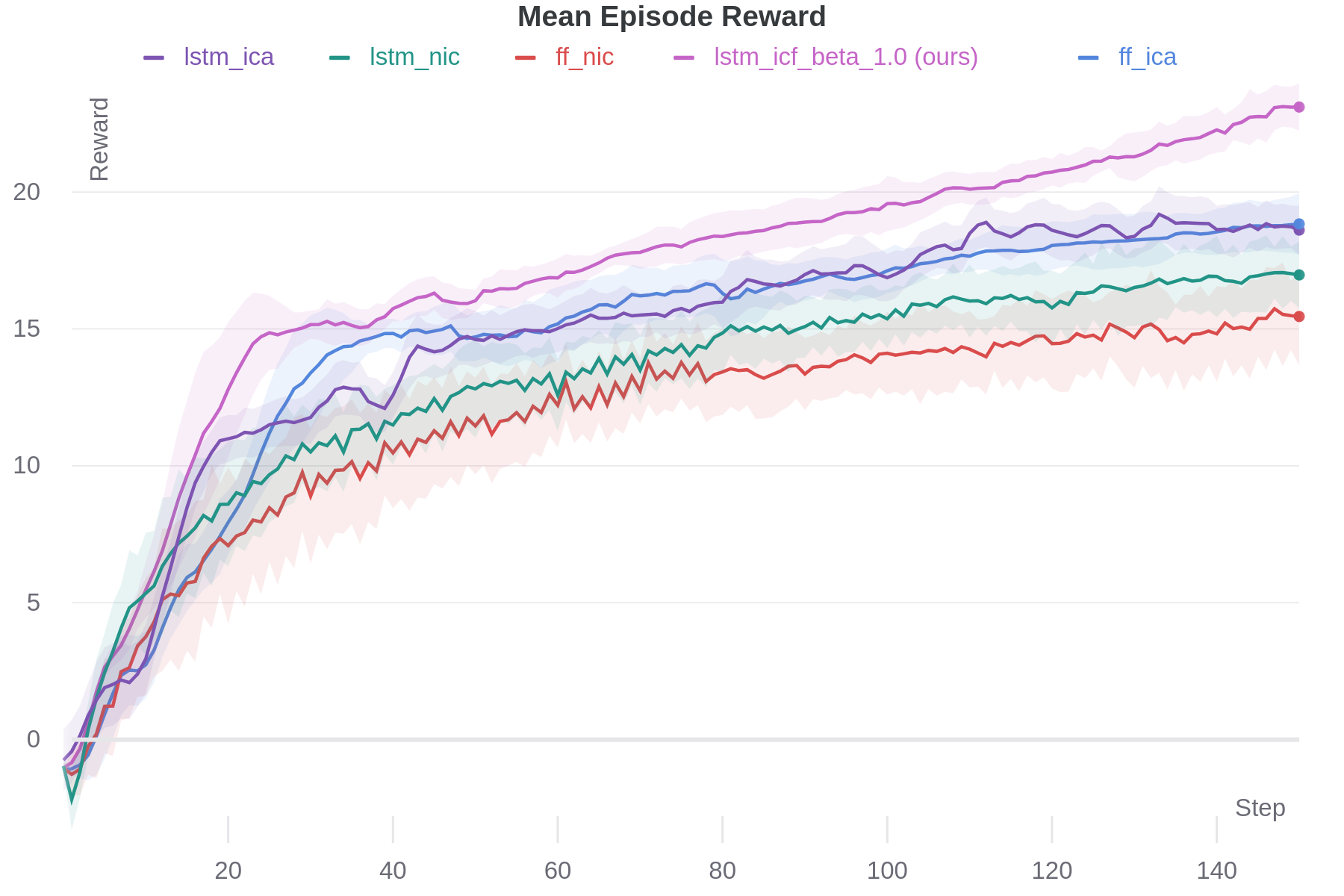}  
        \caption{}
        \label{fig:dd0_runs}   
    \end{subfigure}
    \begin{subfigure}{\textwidth}
        \centering
        \includegraphics[width=0.8\linewidth]{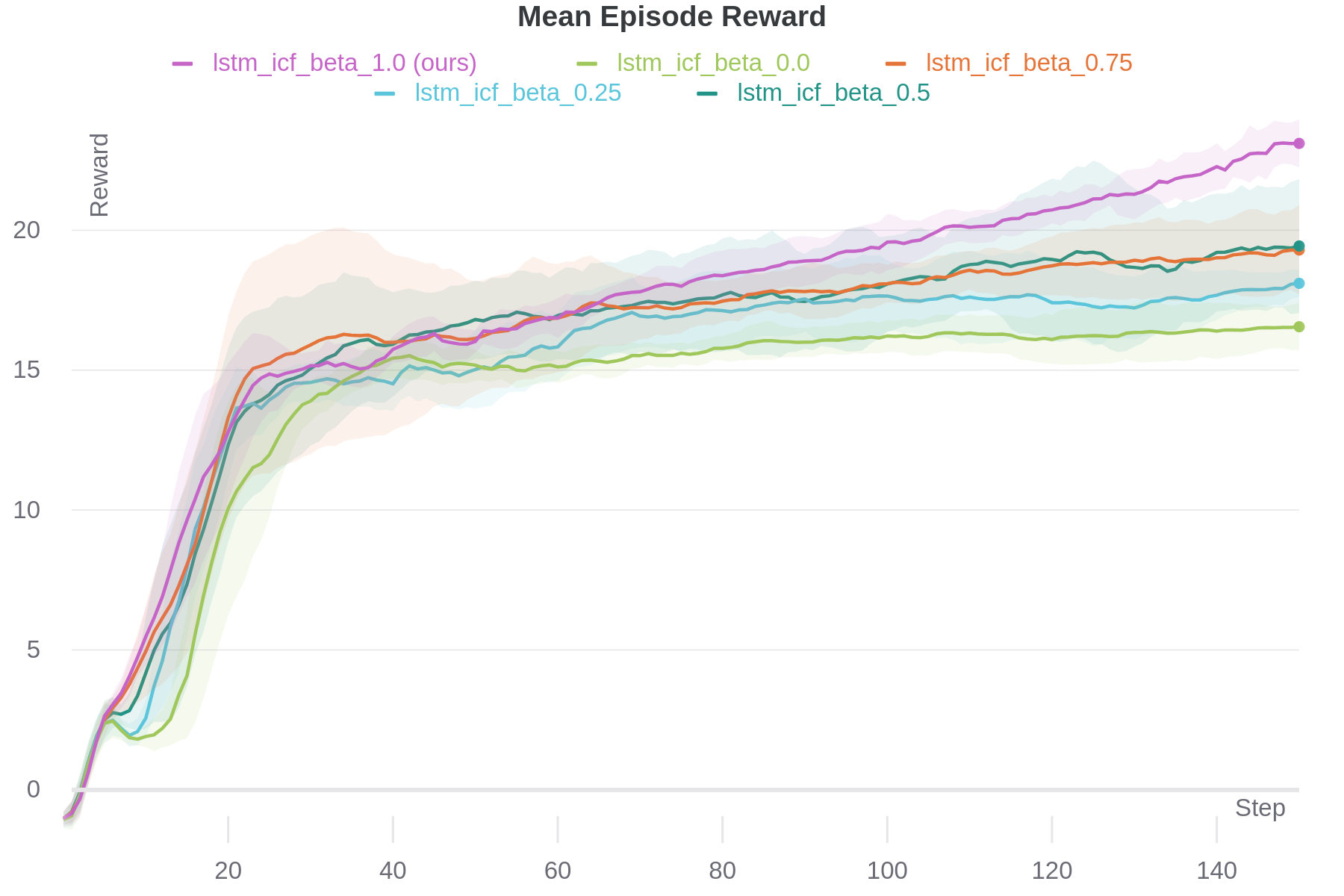}  
        \caption{}
        \label{fig:dd0_beta}
    \end{subfigure}
    \caption{Simulation results in DeepDrive-Zero environment. (a) Mean episode reward for different ablations, (b) mean episode reward for different $\beta$ values. The shaded area shows one standard deviation.}
\end{figure}

\paragraph{DeepDrive-Zero Environment:}
We ran all ablations for ten random seeds in the DeepDrive-Zero environment to test our proposed method. We used self-play in simulations and used the latest set of parameters for actors in each episode.
The results are shown in Fig~\ref{fig:dd0_runs}. The x-axis shows the number of training iterations. In each iteration, we ran 100 parallel environments for 3000 steps and collected data. Next, we updated actors and critic networks using the collected data. After each iteration, we ran the agents for 100 episodes, took the mean of these episodes' rewards (sum of all agents' rewards), and plotted them. The shaded area shows one standard deviation of episode rewards.
The hyperparameters used in the MACRPO algorithm are listed in Table~\ref{tab:macppo_params} in Appendix~\ref{hypers}.

The proposed algorithm, LSTM-ICF, outperforms the ablations. The next best performances are for LSTM-ICA and FF-ICA, which are almost the same. Moreover, information integration in the advantage function, in both FF-ICA and LSTM-ICA, improves the performance compared to FF-NIC and LSTM-NIC; however, the achieved performance gain in the fully connected case is higher. The FF-ICA surpasses LSTM-NIC, which shows the effectiveness of sharing information across agents through the proposed advantage function, even without an LSTM layer. Furthermore, the addition of LSTM layer to add another level of information integration, LSTM-ICF, boosts performance when compared to FF-ICA.
Fig~\ref{fig:dd0_beta} shows the analysis of the effect of different $\beta$ values in Equations~(\ref{eq:ma_discounted_return}, \ref{eq:ma_mean_r}, \ref{eq:ma_delta}). The best performance is for $\beta = 1$, and as the value of $\beta$ is reduced, the agents' performance decreases. We demonstrate the effect of different $\beta$ values in this environment, but for other environments, the results will be provided for $\beta \in \{0, 1\}$ only.

To achieve smooth driving performance, a curriculum-based learning method and gradual weight increase of reward factors were used. The weights of Jerk, G-force, steering angle change, acceleration change, and going out of the lane in the reward function were gradually increased to $3.3 \times 10^{-6}$, $0.1$, $3$, $0.05$, and $0.3$, respectively. We then added termination of episodes for lane violation to force cars to stay between the lanes. 
After curriculum learning and smoothing the driving behavior, the cars follow the waypoints to reach their destination. The car that starts from the bottom and wants to make a left-turn yields nicely for the other agent if they reach the intersection simultaneously and then make the left-turn, and if it has time to cross the intersection before the other agent arrives, it does. A video of the final result can be found in the supplementary materials.

\paragraph{Multi-Walker Environment:}
We ran 20 parallel environments and 2500 time-steps during each update iteration for the Multi-Walker environment. After each iteration, we ran agents for 100 episodes and plotted the mean of these episodes' rewards. Each episode's reward is the sum of all the agents' rewards. Ten different random seeds are used for each ablation. We also used the latest set of parameters for all actors.
The hyperparameters used in the MACRPO algorithm are listed in Table~\ref{tab:macppo_params} in Appendix~\ref{hypers}.

Fig~\ref{fig:mw_runs} shows a massive performance improvement of our proposed method, LSTM-ICF with $\beta = 1$, when compared to ablations. 
LSTM-ICF with $\beta = 0$, information integration through only the LSTM layer, has the next best performance. After these two, LSTM-ICA, which does the information integration using the advantage estimation function, performs better than FF-ICA, FF-NIC, and LSTM-NIC cases.  
The effect of $\beta$ value and information sharing through the advantage estimation function in performance improvement can be seen as we move from LSTM-ICF with $\beta = 0$ to LSTM-ICF with $\beta = 1$ and from FF-NIC to FF-ICA. By comparing FF-ICA and LSTM-ICF, we can also see the impact of information integration using the LSTM layer. Note that the $\beta$ value in FF-ICA is equal to 1.
A video of the trained model can be found in the supplementary materials.

\begin{figure}[t!]
    \includegraphics[width=0.8\linewidth]{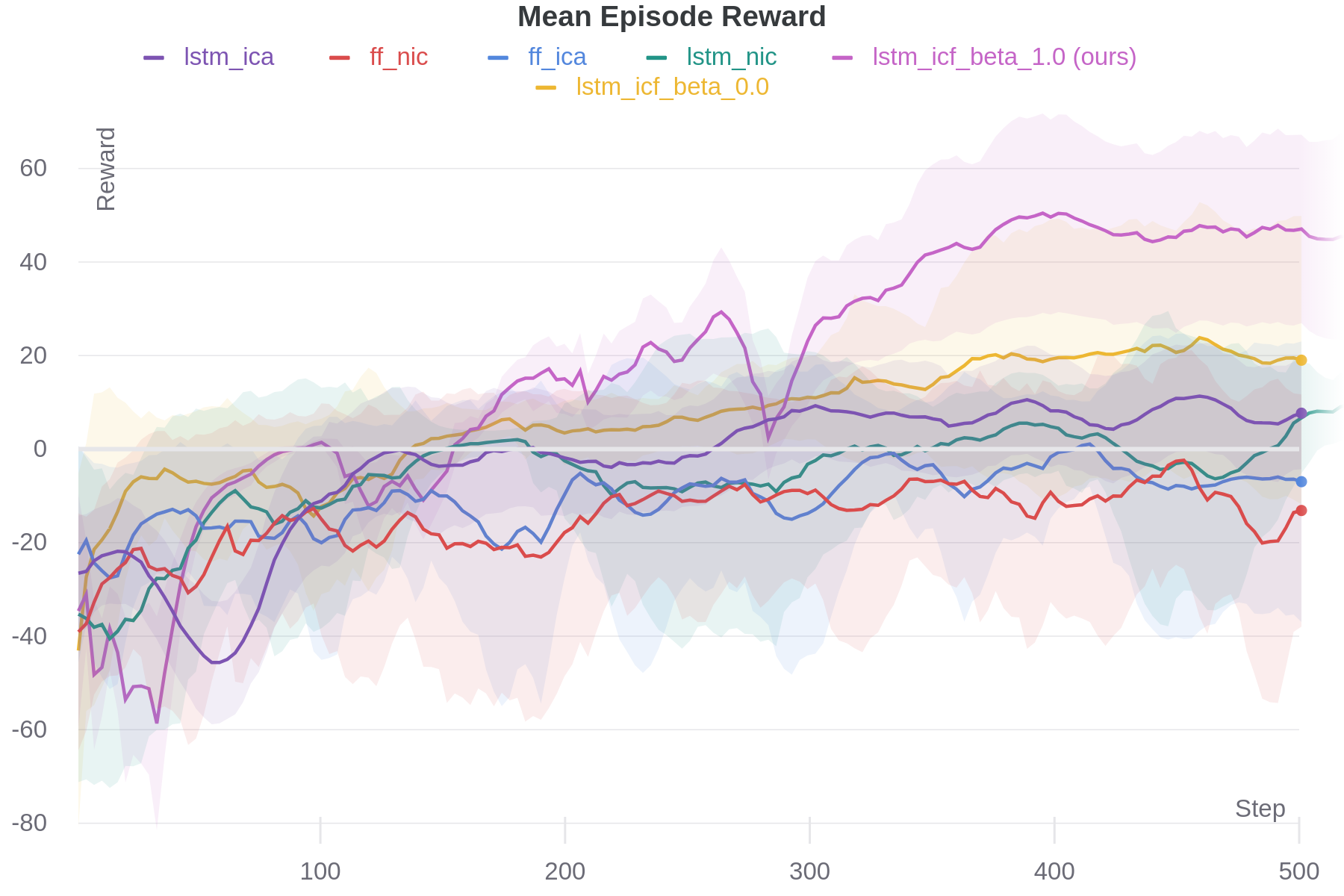}
    \centering
    \caption{Multi-Walker simulation results for different ablations.}
    \label{fig:mw_runs}
\end{figure}

\begin{figure}[h]
    \includegraphics[width=0.8\linewidth]{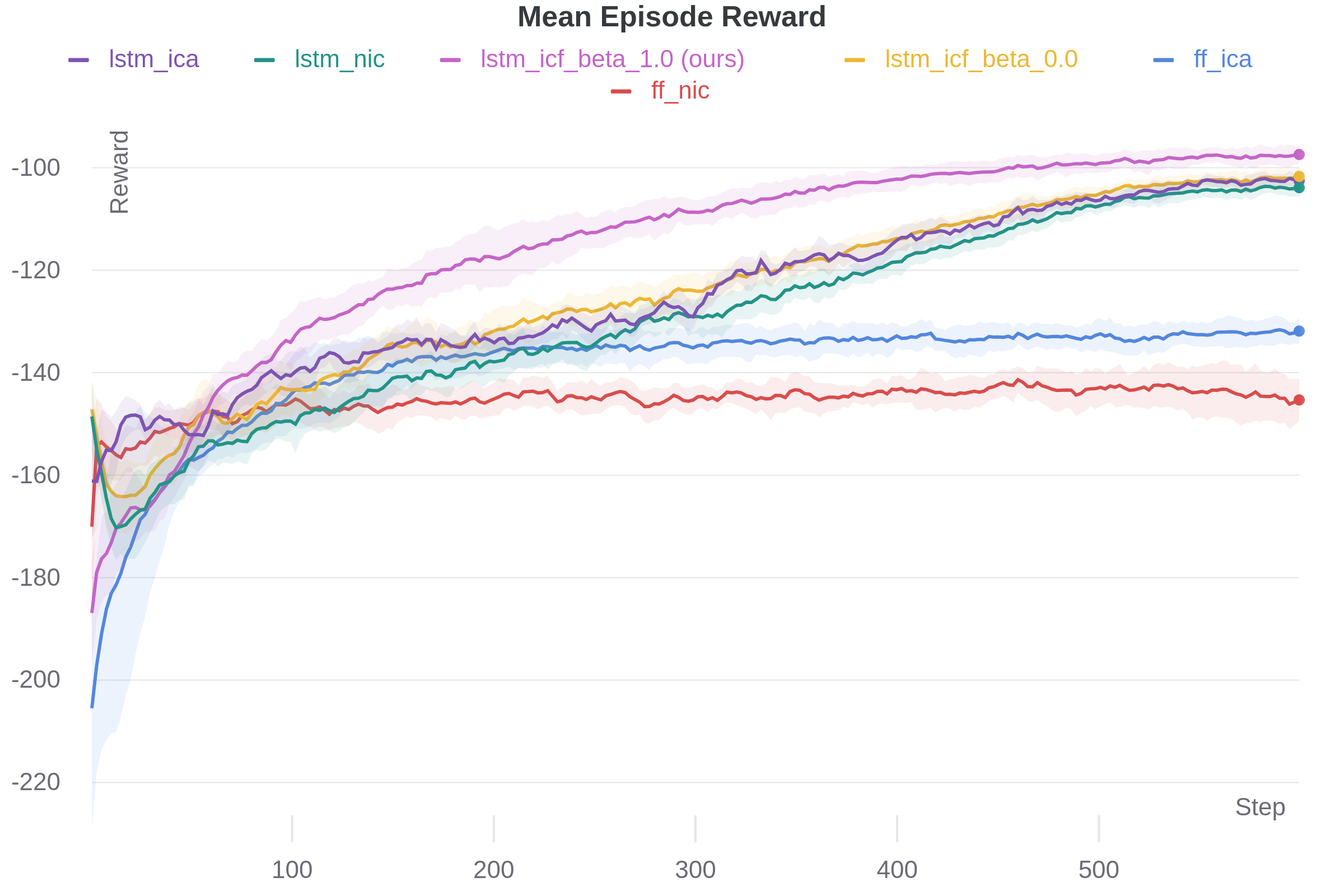}
    \centering
    \caption{Particle environment simulation results for different ablations.}
    \label{fig:particle_runs}
\end{figure}

\paragraph{Cooperative Navigation in Particle Environment:}
In the particle environment, in each iteration, we ran 20 parallel environments to collect data for 2500 time steps and used that data to update the network. The agents were then evaluated using the trained weights for 100 episodes. We ran the simulation with six random seeds. MACRPO hyperparameters are shown in Table~\ref{tab:macppo_params} in Appendix~\ref{hypers}.

The results of this environment are depicted in~\ref{fig:particle_runs}. Similar to the other two environments, the proposed LSTM-ICF with $\beta = 1$ outperforms ablations. The next best performance is achieved with LSTM-ICF with $\beta = 0$, which only uses the LSTM layer that was trained using the created meta-trajectory. Moreover, the LSTM-ICA's performance is almost identical to LSTM-ICF when $\beta = 0$. This shows that both novel ideas cause the same performance gain over LSTM-NIC.
These results show that cases with LSTM layer perform better than feed-forward ones, even in the FF-ICA case, which integrates information through the advantage function.
A video of the trained model can be found in the supplementary materials.

Both ideas were evaluated in the ablation study, and the results clearly demonstrate the effect of the proposed ideas in performance improvement. Ablation studies provide evidence that the findings are not spurious, but are associated with the proposed enhancements. 
$\beta = 1$ corresponds to the total reward over all agents, the optimization goal. However, it is known that such a team reward causes a credit assignment problem since each agent's contribution to the team reward could differ. Due to this, we wanted to experimentally study whether beta values less than one would alleviate the credit assignment problem to the extent that the suboptimality of the reward would be overcome. According to the results of the experiment, this wasn't the case, and $\beta = 1$ gave the best performance. 

Moreover, as the results illustrate, both proposed ideas result in a performance gain, but this is not the same for all environments. In the DeepDrive-Zero environment, information integration through advantage function estimation improves the performance slightly more than the LSTM layer. However, in the Multi-Walker environment, the LSTM layer is more effective, and in the Particle environment, their effect is almost the same.

\subsection{Comparison to State-of-the-Art Methods}

\begin{table}[t!]
    \centering
  \caption{Comparing performance of our method with state-of-the-art approaches. Numbers show max average reward in each environment for ten random seeds, except for the Multi-Walker environment which is 1000 random seeds.}
  \label{tab:sota}
  \begin{tabular}{cccc}\toprule
    \textit{Method} & \rotatebox[origin=c]{0}{\textit{\begin{tabular}{@{}c@{}} DeepDrive- \\ Zero\end{tabular}}} & \rotatebox[origin=c]{0}{\textit{Multi-Walker}} & \rotatebox[origin=c]{0}{\textit{Particle}} \\ \midrule
    DQN & 4 & -100000 & -151.8 \\
    RDQN & 6 & -100000 & 153.2 \\
    A2C & 0.5 & -27.6 & -148.6 \\
    DDPG & 2 & -57.8 & - \\
    PPO & 16 & 41 & -144.3 \\
    SAC & -1.5 & -16.9 & -143.7 \\
    TD3 & -1 & -8 & - \\
    APEX-DQN & 8 & -100000 & -136.2 \\
    APEX-DDPG & 14 & -23 & - \\
    IMPALA & -0.66 & -88 & -155.2 \\
    MADDPG & -0.1 & -96 & -98.3 \\
    QMIX & -0.9 & -24 & -155.6 \\
    Ours ($\beta = 0$) & 17.3 & 24.2 & -100.7 \\
    Ours (full model) & \textbf{23.7} & \textbf{47.8} & \textbf{-95.8} \\ \bottomrule
  \end{tabular}
\end{table}

We compared the proposed method with several state-of-the-art algorithms in each environment. Our method is compared against several single-agent baselines with shared parameters across agents (DQN, RDQN, A2C, DDPG, PPO, SAC, TD3, APEX-DQN, APEX-DDPG, and IMPALA), which were tested in~\cite{terry2020parameter}. We also compared our method to state-of-the-art multi-agent approaches MADDPG~\cite{lowe2017multi} and QMIX~\cite{rashid2018qmix}. 

\begin{figure}[t!]
    \begin{subfigure}{\linewidth}
        \includegraphics[width=0.8\linewidth]{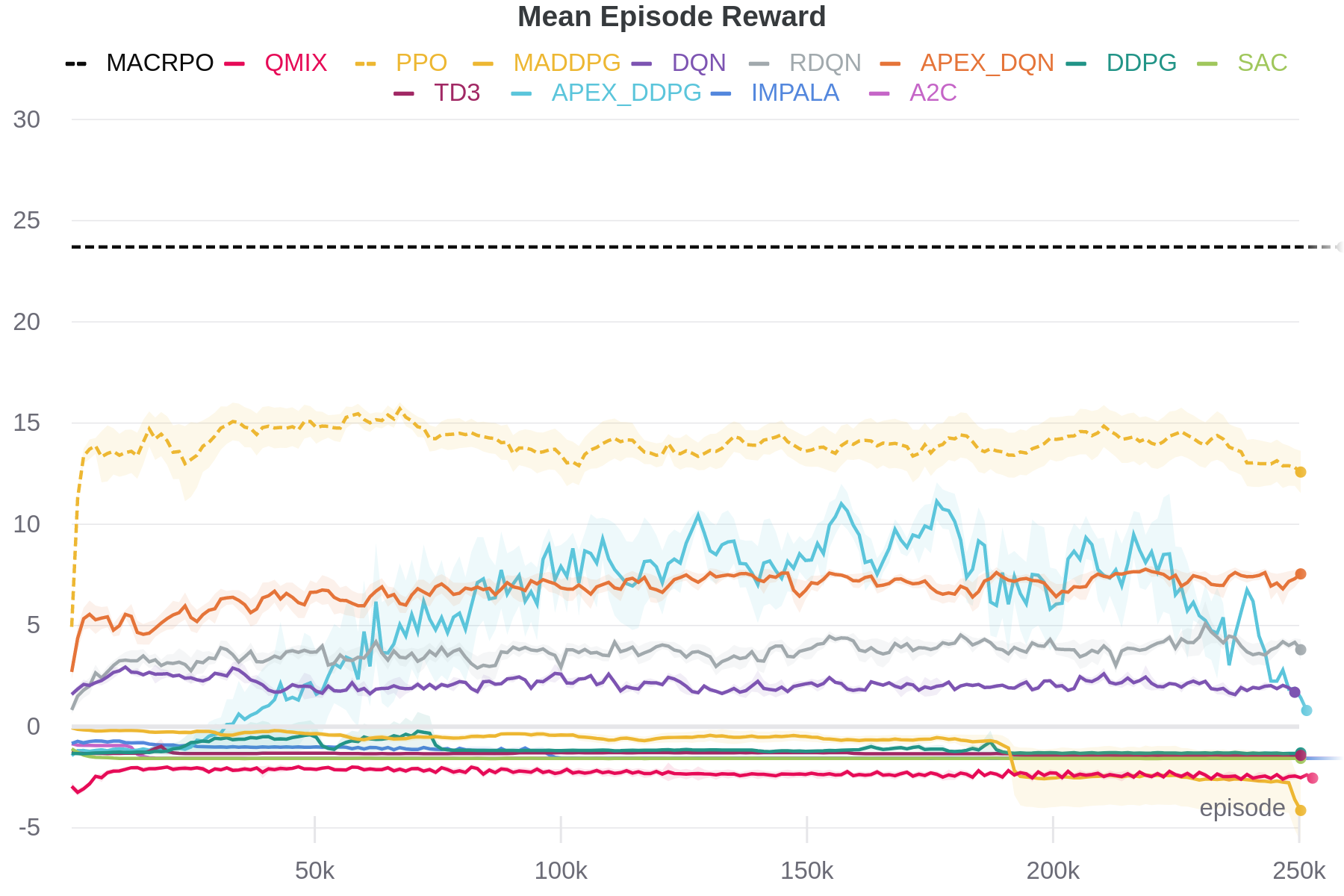}
        \centering
        \caption{}
    \end{subfigure}
    
    \begin{subfigure}{\linewidth}
        \centering
        \includegraphics[width=0.8\linewidth]{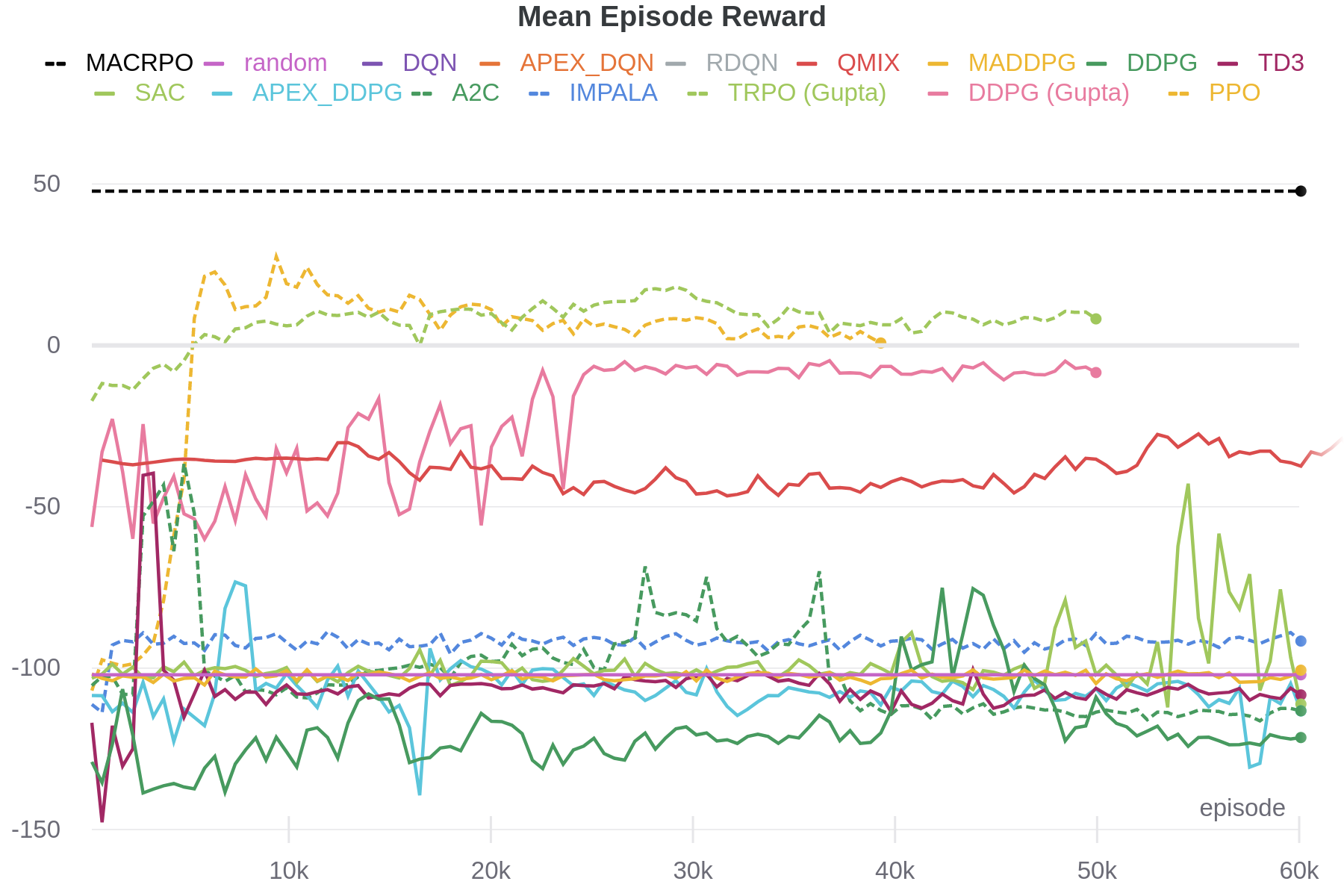}  
        \caption{}
    \end{subfigure}
    \begin{subfigure}{\linewidth}
        \centering
        \includegraphics[width=0.8\linewidth]{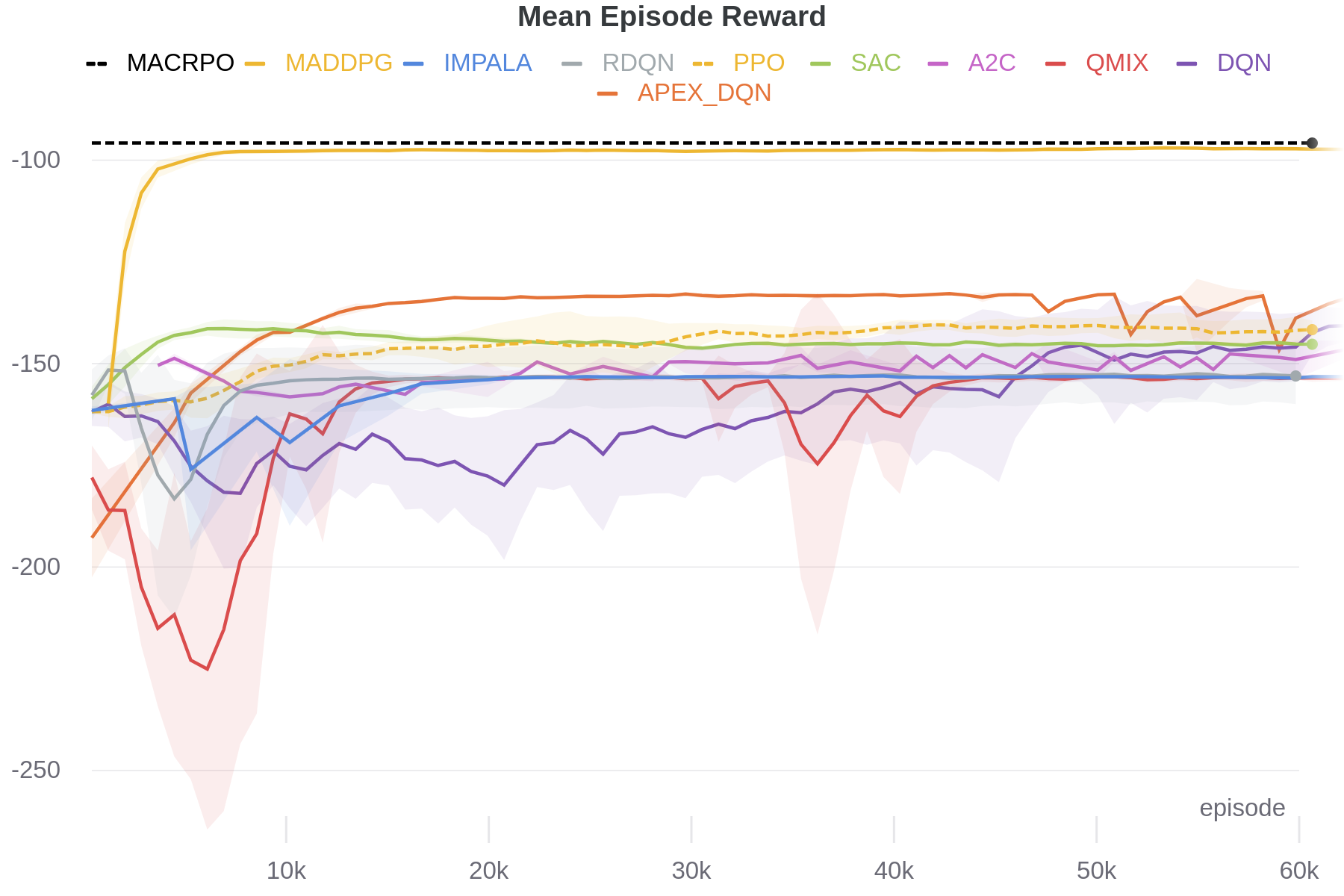}  
        \caption{}
    \end{subfigure}
    \caption{Analysis of baseline algorithms proposed in~\cite{terry2020parameter} in three environments: (a) DeepDrive-Zero, (b) Multi-Walker, and (c) Particle environments.}
    \label{fig:runs_baselines}
\end{figure}

Each agent's goal in MACRPO is to maximize the total reward of all agents, while the goal of other methods is to maximize the total reward of each agent without considering other agents' reward in their objective function. 
In order to have a more fair comparison, We report the result for our method when $\beta = 0$ too. 
The results are shown in Table~\ref{tab:sota}. 
To get a better idea of the performance of the algorithms, the mean episode reward for different baseline algorithms in three environments are shown in Fig~\ref{fig:runs_baselines}.

\paragraph{DeepDrive-Zero Environment:}
In this environment, our full method and also the case with $\beta = 0$ achieved the most average reward. The next best was PPO with parameter sharing between agents followed by APEX-DQN and APEX-DDPG. An environment with discretized action space was used for algorithms with discretized action space.

\paragraph{Multi-Walker Environment:}
Similar to the previous environment, the proposed method outperformed other methods by a large margin with an average reward of 47.8. Next, PPO with parameter sharing had the second-best performance with a maximum average reward of 41. Our method with $\beta = 0$ achieved the third best average reward. The baselines' results reported for this environment in Table~\ref{tab:sota} are taken from~\cite{terry2020parameter}.

\paragraph{Cooperative Navigation in Particle Environment:}
As in both previous environments, our approach outperformed other approaches in this environment as well, although the difference was minor compared to MADDPG. Our method with $\beta = 0$ is in the third place after MADDPG with small margin.
In this environment with discrete action space, we used a categorical distribution instead of a multivariate Gaussian distribution. Algorithms with continuous action spaces were not tested in this environment, and are marked with a dash in the table. Adapting these algorithms for discrete action environments could be achieved using the same trick, but we did not make changes to the standard implementation for baselines from ~\cite{terry2020parameter}.

For the multi-walker environment, we used the hyperparameters from ~\cite{terry2020parameter}; for other environments, we started from the original parameters and further tuned them using grid search to improve performance. All hyperparameters for each algorithm are included in Appendix~\ref{hypers}.

The results show that the performance benefit given by the two proposed ways of sharing information across agents is significant such that the method outperforms state-of-the-art algorithms.

\clearpage

\section{Conclusion \& Future Work}
\label{conclusion}

In this paper, MACRPO, a centralized training and decentralized execution framework for multi-agent cooperative settings was presented. The framework is applicable to both discrete and continuous action spaces. 
In addition to parameter sharing across agents, this framework integrates information across agents and time in two novel ways: network architecture and the advantage estimation function. 
An ablation study in three environments revealed that both ways of information sharing are beneficial. 
Furthermore, the method was compared to state-of-the-art multi-agent algorithms such as QMIX and MADDPG, as well as single-agent algorithms that share parameters between agents, such as IMPALA and APEX. The results showed that the proposed algorithm performed significantly better than state-of-the-art algorithms.
A single recurrent network to summarize the state of all agents may be problematic when the number of agents is large. A potential solution to this problem could be to use an attention mechanism for the agent to learn on which other agents to pay attention to, warranting further study to realize the potential of the proposed approach with a high number of agents.

\section*{ACKNOWLEDGMENT}

The authors wish to acknowledge CSC – IT Center for Science, Finland, for generous computational resources.
We also acknowledge the computational resources provided by the Aalto Science-IT project.


%
%

\clearpage



\printbibliography

@article{terry2020parameter,
  title={Parameter Sharing is Surprisingly Useful for Multi-Agent Deep Reinforcement Learning},
  author={Terry, Justin K and Grammel, Nathaniel and Hari, Ananth and Santos, Luis and Black, Benjamin and Manocha, Dinesh},
  journal={arXiv preprint arXiv:2005.13625},
  year={2020}
}

@article{zhao2020research,
  title={Research on the Multiagent Joint Proximal Policy Optimization Algorithm Controlling Cooperative Fixed-Wing UAV Obstacle Avoidance},
  author={Zhao, Weiwei and Chu, Hairong and Miao, Xikui and Guo, Lihong and Shen, Honghai and Zhu, Chenhao and Zhang, Feng and Liang, Dongxin},
  journal={Sensors},
  volume={20},
  number={16},
  pages={4546},
  year={2020},
  publisher={Multidisciplinary Digital Publishing Institute}
}

@article{bansal2017emergent,
  title={Emergent complexity via multi-agent competition},
  author={Bansal, Trapit and Pachocki, Jakub and Sidor, Szymon and Sutskever, Ilya and Mordatch, Igor},
  journal={arXiv preprint arXiv:1710.03748},
  year={2017}
}

@article{hernandez2019survey,
  title={A survey and critique of multiagent deep reinforcement learning},
  author={Hernandez-Leal, Pablo and Kartal, Bilal and Taylor, Matthew E},
  journal={Autonomous Agents and Multi-Agent Systems},
  volume={33},
  number={6},
  pages={750--797},
  year={2019},
  publisher={Springer}
}

@article{shalev2016safe,
  title={Safe, multi-agent, reinforcement learning for autonomous driving},
  author={Shalev-Shwartz, Shai and Shammah, Shaked and Shashua, Amnon},
  journal={arXiv preprint arXiv:1610.03295},
  year={2016}
}

@article{berner2019dota,
  title={Dota 2 with large scale deep reinforcement learning},
  author={Berner, Christopher and Brockman, Greg and Chan, Brooke and Cheung, Vicki and D{\k{e}}biak, Przemys{\l}aw and Dennison, Christy and Farhi, David and Fischer, Quirin and Hashme, Shariq and Hesse, Chris and others},
  journal={arXiv preprint arXiv:1912.06680},
  year={2019}
}

@article{vinyals2019grandmaster,
  title={Grandmaster level in StarCraft II using multi-agent reinforcement learning},
  author={Vinyals, Oriol and Babuschkin, Igor and Czarnecki, Wojciech M and Mathieu, Micha{\"e}l and Dudzik, Andrew and Chung, Junyoung and Choi, David H and Powell, Richard and Ewalds, Timo and Georgiev, Petko and others},
  journal={Nature},
  volume={575},
  number={7782},
  pages={350--354},
  year={2019},
  publisher={Nature Publishing Group}
}

@article{ying2005multi,
  title={Multi-agent framework for third party logistics in E-commerce},
  author={Ying, Wang and Dayong, Sang},
  journal={Expert Systems with Applications},
  volume={29},
  number={2},
  pages={431--436},
  year={2005},
  publisher={Elsevier}
}

@inproceedings{tan1993multi,
  title={Multi-agent reinforcement learning: Independent vs. cooperative agents},
  author={Tan, Ming},
  booktitle={Proceedings of the tenth international conference on machine learning},
  pages={330--337},
  year={1993}
}

@article{nguyen2020deep,
  title={Deep reinforcement learning for multiagent systems: A review of challenges, solutions, and applications},
  author={Nguyen, Thanh Thi and Nguyen, Ngoc Duy and Nahavandi, Saeid},
  journal={IEEE transactions on cybernetics},
  year={2020},
  publisher={IEEE}
}

@article{kraemer2016multi,
  title={Multi-agent reinforcement learning as a rehearsal for decentralized planning},
  author={Kraemer, Landon and Banerjee, Bikramjit},
  journal={Neurocomputing},
  volume={190},
  pages={82--94},
  year={2016},
  publisher={Elsevier}
}

@inproceedings{foerster2016learning,
  title={Learning to communicate with deep multi-agent reinforcement learning},
  author={Foerster, Jakob and Assael, Ioannis Alexandros and De Freitas, Nando and Whiteson, Shimon},
  booktitle={Advances in neural information processing systems},
  pages={2137--2145},
  year={2016}
}

@inproceedings{lowe2017multi,
  title={Multi-agent actor-critic for mixed cooperative-competitive environments},
  author={Lowe, Ryan and Wu, Yi I and Tamar, Aviv and Harb, Jean and Abbeel, OpenAI Pieter and Mordatch, Igor},
  booktitle={Advances in neural information processing systems},
  pages={6379--6390},
  year={2017}
}

@article{foerster2017counterfactual,
  title={Counterfactual multi-agent policy gradients},
  author={Foerster, Jakob and Farquhar, Gregory and Afouras, Triantafyllos and Nardelli, Nantas and Whiteson, Shimon},
  journal={arXiv preprint arXiv:1705.08926},
  year={2017}
}

@article{matignon2012independent,
  title={Independent reinforcement learners in cooperative Markov games: a survey regarding coordination problems.},
  author={Matignon, Laetitia and Laurent, Guillaume J and Le Fort-Piat, Nadine},
  year={2012}
}

@inproceedings{gupta2017cooperative,
  title={Cooperative multi-agent control using deep reinforcement learning},
  author={Gupta, Jayesh K and Egorov, Maxim and Kochenderfer, Mykel},
  booktitle={International Conference on Autonomous Agents and Multiagent Systems},
  pages={66--83},
  year={2017},
  organization={Springer}
}

@inproceedings{sunehag2018value,
  title={Value-Decomposition Networks For Cooperative Multi-Agent Learning Based On Team Reward.},
  author={Sunehag, Peter and Lever, Guy and Gruslys, Audrunas and Czarnecki, Wojciech Marian and Zambaldi, Vin{\'\i}cius Flores and Jaderberg, Max and Lanctot, Marc and Sonnerat, Nicolas and Leibo, Joel Z and Tuyls, Karl and others},
  booktitle={AAMAS},
  pages={2085--2087},
  year={2018}
}

@article{rashid2018qmix,
  title={QMIX: Monotonic value function factorisation for deep multi-agent reinforcement learning},
  author={Rashid, Tabish and Samvelyan, Mikayel and De Witt, Christian Schroeder and Farquhar, Gregory and Foerster, Jakob and Whiteson, Shimon},
  journal={arXiv preprint arXiv:1803.11485},
  year={2018}
}

@article{wang2020r,
  title={R-maddpg for partially observable environments and limited communication},
  author={Wang, Rose E and Everett, Michael and How, Jonathan P},
  journal={arXiv preprint arXiv:2002.06684},
  year={2020}
}

@incollection{littman1994markov,
  title={Markov games as a framework for multi-agent reinforcement learning},
  author={Littman, Michael L},
  booktitle={Machine learning proceedings 1994},
  pages={157--163},
  year={1994},
  publisher={Elsevier}
}

@article{schulman2017proximal,
  title={Proximal policy optimization algorithms},
  author={Schulman, John and Wolski, Filip and Dhariwal, Prafulla and Radford, Alec and Klimov, Oleg},
  journal={arXiv preprint arXiv:1707.06347},
  year={2017}
}

@software{craig_quiter_2020_3871907,
  author       = {Craig Quiter},
  title        = {Deepdrive Zero},
  month        = jun,
  year         = 2020,
  publisher    = {Zenodo},
  version      = {alpha},
  doi          = {10.5281/zenodo.3871907},
  url          = {https://doi.org/10.5281/zenodo.3871907}
}

@misc{pettingZoo2020,
  author = {Terry, Justin K and Black, Benjamin and Jayakumar, Mario  and Hari, Ananth and Santos, Luis and Dieffendahl, Clemens and Williams, Niall and Ravi, Praveen and Lokesh, Yashas and Horsch, Caroline and Patel, Dipam},
  title = {Petting{Z}oo},
  year = {2020},
  publisher = {GitHub},
  note = {GitHub repository},
  howpublished = {\url{https://github.com/PettingZoo-Team/PettingZoo}}
}

@article{mordatch2017emergence,
  title={Emergence of Grounded Compositional Language in Multi-Agent Populations},
  author={Mordatch, Igor and Abbeel, Pieter},
  journal={arXiv preprint arXiv:1703.04908},
  year={2017}
}

@article{mao2020learning,
  title={Learning multi-agent communication with double attentional deep reinforcement learning},
  author={Mao, Hangyu and Zhang, Zhengchao and Xiao, Zhen and Gong, Zhibo and Ni, Yan},
  journal={Autonomous Agents and Multi-Agent Systems},
  volume={34},
  number={1},
  pages={1--34},
  year={2020},
  publisher={Springer}
}

@article{kullu2017acmics,
  title={ACMICS: an agent communication model for interacting crowd simulation},
  author={Kullu, Kurtulus and G{\"u}d{\"u}kbay, U{\u{g}}ur and Manocha, Dinesh},
  journal={Autonomous Agents and Multi-Agent Systems},
  volume={31},
  number={6},
  pages={1403--1423},
  year={2017},
  publisher={Springer}
}

\appendix
\clearpage

\section{Hyperparameters}\label{hypers}
Hyperparameters used in MACRPO for three environments are described in Table~\ref{tab:macppo_params}.

\begin{table}[h!]
\centering
  \caption{MACRPO hyperparameters for three MARL environments}
  \label{tab:macppo_params}
  \begin{tabular}{cccc}\toprule
    \textit{Param.} & \rotatebox[origin=c]{60}{\textit{DeepDrive-Zero}} & \rotatebox[origin=c]{60}{\textit{Multi-Walker}} & \rotatebox[origin=c]{60}{\textit{Particle}} \\ \midrule
    actor hidden size & 64 & 32 & 128 \\
    critic hidden size & 128 & 32 & 128 \\
    batch size & 512 & 32 & 1500 \\
    discount & 0.99 & 0.99 & 0.99 \\
    GAE lambda & 0.94 & 0.95 &0.95 \\
    PPO clip & 0.15 & 0.3 & 0.2 \\
    PPO epochs & 4 & 4 & 10 \\
    max grad norm & 1.0 & 1.0 & 1.0  \\
    entropy factor & 0.001 & 0.01 & 0.01 \\
    learning rate & 0.0002 & 0.001 & 0.005 \\
    \begin{tabular}{@{}c@{}}recurrent sequence \\ length (time-step)\end{tabular}  & 20 & 40 & 3 \\
    no. of recurrent layers & 1 & 1 & 1 \\ \bottomrule
  \end{tabular}
\end{table}

The architecture and hyperparameters used for other baselines are taken from~\cite{terry2020parameter} with some fine-tuning to get better performance, and are shown in Tables~\ref{tab:baseline_params1},~\ref{tab:baseline_params2}, and~\ref{tab:baseline_params3}. Some hyperparameter values are constant across all RL methods for all environments. These constant values are reported in \autoref{tab:baseline_params4}. We used the source code for all algorithms from~\cite{terry2020parameter} except for MADDPG which we used the original implementation~\cite{lowe2017multi}.

\begin{table*}[h!b]
    \small
    \caption{Hyperparameters for three MARL environments}
    \centering
    \begin{tabular}{c c c c c}
    \toprule
    RL method & Hyperparameter & DeepDrive-Zero & Multi-Walker & Particle \\
    \midrule
    PPO & \texttt{sample\_batch\_size} & 100 & 100 & 100 \\
     & \texttt{train\_batch\_size} & 5000 & 5000 & 5000 \\
     & \texttt{sgd\_minibatch\_size} & 500 & 500 & 1000 \\
     & \texttt{lambda} & 0.95 & 0.95 & 0.95 \\
     & \texttt{kl\_coeff} & 0.5 & 0.5 & 0.5 \\
     & \texttt{entropy\_coeff} & 0.01 & 0.01 & 0.001 \\
     & \texttt{num\_sgd\_iter} & 10 & 10 & 50 \\
     & \texttt{vf\_clip\_param} & 10.0 & 10.0 & 1.0 \\
     & \texttt{clip\_param} & 0.1 & 0.1 & 0.5 \\
     & \texttt{vf\_share\_layers} & \texttt{True} & \texttt{True} & \texttt{True} \\
     & \texttt{clip\_rewards} & \texttt{True} & \texttt{True} & \texttt{False} \\
     & \texttt{batch\_mode} & \texttt{truncate\_episodes} & \texttt{truncate\_episodes} & \texttt{truncate\_episodes} \\
    \midrule
    IMPALA & \texttt{sample\_batch\_size} & 20 & 20 & 20 \\
     & \texttt{train\_batch\_size} & 512 & 512 & 512 \\
     & \texttt{lr\_schedule} & [[0, 5e-3], [2e7, 1e-12]] & [[0, 5e-3], [2e7, 1e-12]] & [[0, 5e-3], [2e7, 1e-12]] \\
     & \texttt{clip\_rewards} & \texttt{True} & \texttt{True} & \texttt{False} \\
    \midrule
    A2C & \texttt{sample\_batch\_size} & 20 & 20 & 20 \\
     & \texttt{train\_batch\_size} & 512 & 512 & 512 \\
     & \texttt{lr\_schedule} & [[0, 7e-3], [2e7, 1e-12]] & [[0, 7e-3], [2e7, 1e-12]] & [[0, 7e-3], [2e7, 1e-12]] \\
    \midrule
    SAC & \texttt{sample\_batch\_size} & 20 & 20 & 20 \\
     & \texttt{train\_batch\_size} & 512 & 512 & 512 \\
     & \texttt{Q\_model} 
     & \{\texttt{activation:} 
     & \{\texttt{activation:} 
     & \{\texttt{activation:}  \\
     & &\texttt{relu}, 
     & \texttt{relu}, 
     & \texttt{relu}, \\
     & &\texttt{layer\_sizes}: 
     & \texttt{layer\_sizes}: 
     & \texttt{layer\_sizes}: \\
     & &\texttt{[266, 256]}\} 
     & \texttt{[266, 256]}\} 
     & \texttt{[266, 256]}\} \\
     & \texttt{optimization} 
     &  \{\texttt{actor\_lr}: 
     & \{\texttt{actor\_lr}:
     & \{\texttt{actor\_lr}: \\
     & &  0.0003, &  0.0003, &  0.0003, \\
     & &\texttt{actor\_lr}: 
     &\texttt{actor\_lr}: 
     &\texttt{actor\_lr}:\\
     & & 0.0003, & 0.0003, & 0.0003, \\
     & &\texttt{entropy\_lr}: 
     &\texttt{entropy\_lr}: 
     &\texttt{entropy\_lr}: \\
     & & 0.0003,\} & 0.0003,\} & 0.0003,\} \\
     & \texttt{clip\_actions} & \texttt{False} & \texttt{False} & \texttt{False} \\
     & \texttt{exploration\_enabled} & \texttt{True} & \texttt{True} & \texttt{True} \\
     & \texttt{no\_done\_at\_end} & \texttt{True} & \texttt{True} & \texttt{True} \\
     & \texttt{normalize\_actions} & \texttt{False} & \texttt{False} & \texttt{False} \\
     & \texttt{prioritized\_replay} & \texttt{False} & \texttt{False} & \texttt{False} \\
     & \texttt{soft\_horizon} & \texttt{False} & \texttt{False} & \texttt{False} \\
     & \texttt{target\_entropy} & \texttt{auto} & \texttt{auto} & \texttt{auto} \\
     & \texttt{tau} & 0.005 & 0.005 & 0.005 \\
     & \texttt{n\_step} & 1 & 1 & 5 \\
     & \begin{tabular}{@{}c@{}}\texttt{evaluation\_} \\ \texttt{interval}\end{tabular} & 1 & 1 & 1 \\
     & \begin{tabular}{@{}c@{}}\texttt{metrics\_smoothing\_} \\ \texttt{episodes}\end{tabular} & 5 & 5 & 5 \\
     & \begin{tabular}{@{}c@{}}\texttt{target\_network\_} \\ \texttt{update\_freq}\end{tabular} & 1 & 1 & 1 \\
     & \texttt{learning\_starts} & 1000 & 1000 & 1000 \\
     & \begin{tabular}{@{}c@{}}\texttt{timesteps\_per\_} \\ \texttt{iteration}\end{tabular} & 1000 & 1000 & 1000 \\
     & \texttt{buffer\_size} & 100000 & 100000 & 100000 \\
    \bottomrule
    \end{tabular}
    \label{tab:baseline_params1}
\end{table*}

\begin{table*}[ht]
\small
\caption{Hyperparameters for DeepDrive-Zero, Multi-Walker, and Particle environments}
\centering
    \begin{tabular}{c c c c c}
    \toprule
    RL method & Hyperparameter & DeepDrive-Zero & Multi-Walker & Particle \\
    \midrule
    APEX-DQN & \texttt{sample\_batch\_size} & 20 & 20 & 20 \\
     & \texttt{train\_batch\_size} & 32 & 512 & 5000 \\
     & \texttt{learning\_starts} & 1000 & 1000 & 1000 \\
     & \texttt{buffer\_size} & 100000 & 100000 & 100000 \\
     & \texttt{dueling} & \texttt{True} & \texttt{True} & \texttt{True} \\
     & \texttt{double\_q} & \texttt{True} & \texttt{True} & \texttt{True} \\
    \midrule
    Rainbow-DQN & \texttt{sample\_batch\_size} & 20 & 20 & 20 \\
     & \texttt{train\_batch\_size} & 32 & 512 & 1000 \\
     & \texttt{learning\_starts} & 1000 & 1000 & 1000 \\
     & \texttt{buffer\_size} & 100000 & 100000 & 100000 \\
     & \texttt{n\_step} & 2 & 2 & 2 \\
     & \texttt{num\_atoms} & 51 & 51 & 51 \\
     & \texttt{v\_min} & 0 & 0 & 0 \\
     & \texttt{v\_max} & 1500 & 1500 & 1500 \\
     & \texttt{prioritized\_replay} & \texttt{True} & \texttt{True} & \texttt{True} \\
     & \texttt{dueling} & \texttt{True} & \texttt{True} & \texttt{True} \\
     & \texttt{double\_q} & \texttt{True} & \texttt{True} & \texttt{True} \\
     & \texttt{parameter\_noise} & \texttt{True} & \texttt{True} & \texttt{True} \\
     & \texttt{batch\_mode} & \texttt{complete\_episodes} & \texttt{complete\_episodes} & \texttt{complete\_episodes}\\
    \midrule
    Plain DQN & \texttt{sample\_batch\_size} & 20 & 20 & 20 \\
     & \texttt{train\_batch\_size} & 32 & 512 & 5000 \\
     & \texttt{learning\_starts} & 1000 & 1000 & 1000 \\
     & \texttt{buffer\_size} & 100000 & 100000 & 100000 \\
     & \texttt{dueling} & \texttt{False} & \texttt{False} & \texttt{False} \\
     & \texttt{double\_q} & \texttt{False} & \texttt{False} & \texttt{False} \\
    \midrule
    QMIX & \texttt{buffer\_size} & 10000 & 3000 & 100000 \\
     & \texttt{gamma} & 0.99 & 0.99 & 0.99 \\
     & \texttt{critic\_lr} & 0.001 & 0.0005 & 0.001 \\
     & \texttt{lr} & 0.001 & 0.0005 & 0.001 \\
     & \texttt{grad\_norm\_clip} & 10 & 10 & 10 \\
     & \texttt{optim\_alpha} & 0.99 & 0.99 & 0.99 \\
     & \texttt{optim\_eps} & 0.00001 & 0.05 & 0.00001 \\
     & \texttt{epsilon\_finish} & 0.02 & 0.05 & 0.02 \\
     & \texttt{epsilon\_start} & 1.0 & 1.0 & 1.0 \\
    \midrule
    MADDPG & \texttt{lr} & 0.001 & 0.0001 & 0.01 \\
     & \texttt{batch\_size} & 64 & 512 & 500 \\
     & \texttt{num\_envs} & 1 & 64 & 1 \\
     & \texttt{num\_cpus} & 1 & 8 & 1 \\
     & \texttt{buffer\_size} & 1e5 & 1e5 & 1e5 \\
     & \texttt{steps\_per\_update} & 4 & 4 & 4 \\
     \bottomrule
    \end{tabular}
    \label{tab:baseline_params2}
\end{table*}

\begin{table*}[ht]
\small
\caption{Hyperparameters for DeepDrive-Zero and Multi-Walker}
\centering
    \begin{tabular}{c c c c}
    \toprule
     RL method & Hyperparameter & DeepDrive-Zero & Multi-Walker \\
    \midrule
    APEX-DDPG & \texttt{sample\_batch\_size} & 20 & 20 \\
     & \texttt{train\_batch\_size} & 512 & 512 \\
     & \texttt{lr} & 0.0001 & 0.0001 \\
     & \texttt{beta\_annealing\_fraction} & 1.0 & 1.0 \\
     & \texttt{exploration\_fraction} & 0.1 & 0.1 \\
     & \texttt{final\_prioritized\_replay\_beta} & 1.0 & 1.0 \\
     & \texttt{n\_step} & 3 & 3 \\
     & \texttt{prioritized\_replay\_alpha} & 0.5 & 0.5 \\
     & \texttt{learning\_starts} & 1000 & 1000 \\
     & \texttt{buffer\_size} & 100000 & 100000 \\
     & \texttt{target\_network\_update\_freq} & 50000 & 50000 \\
     & \texttt{timesteps\_per\_iteration} & 2500 & 25000\\
    \midrule
    Plain DDPG & \texttt{sample\_batch\_size} & 20 & 20 \\
     & \texttt{train\_batch\_size} & 512 & 512 \\
     & \texttt{learning\_starts} & 5000 & 5000 \\
     & \texttt{buffer\_size} & 100000 & 100000 \\
     & \texttt{critics\_hidden} & [256, 256] & [256, 256] \\
    \midrule
    TD3 & \texttt{sample\_batch\_size} & 20 & 20 \\
     & \texttt{train\_batch\_size} & 512 & 512 \\
     & \texttt{critics\_hidden} & [256, 256] & [256, 256] \\
     & \texttt{learning\_starts} & 5000 & 5000 \\
     & \texttt{pure\_exploration\_steps} & 5000 & 5000 \\
     & \texttt{buffer\_size} & 100000 & 100000 \\
    \bottomrule
    \end{tabular}
    \label{tab:baseline_params3}
\end{table*}



\begin{table*}[ht]
\small
\caption{Variables set to constant values across all RL methods for all environments}
    \centering
    \begin{tabular}{c c}
        \toprule
        Variable & Value set in all RL methods\\
        \midrule
        \# worker threads & 8\\
        \# envs per worker & 8\\
        \texttt{gamma} & 0.99 \\
        MLP hidden layers & [400, 300]\\
        \bottomrule
    \end{tabular}
    \label{tab:baseline_params4}
\end{table*}

\end{document}